\def\isarxiv{1} 

\ifdefined\isarxiv
\documentclass[11pt]{article}

\usepackage[numbers]{natbib}

\else
\documentclass{article}
\usepackage{neurips_2023}
\fi

\usepackage{amsmath}
\usepackage{amsthm}
\usepackage{amssymb}
\usepackage{algorithm}
\usepackage{subfig}
\usepackage{algpseudocode}
\usepackage{graphicx}
\usepackage{grffile}
\usepackage{wrapfig,epsfig}
\usepackage{url}
\usepackage{xcolor}
\usepackage{epstopdf}

\usepackage{bbm}
\usepackage{dsfont}

\allowdisplaybreaks

\ifdefined\isarxiv

\usepackage{tikz}
\usepackage{hyperref}  
\hypersetup{colorlinks=true,citecolor=blue,linkcolor=blue} 
\usetikzlibrary{arrows}
\usepackage[margin=1in]{geometry}

\else

\usepackage{microtype}
\usepackage{hyperref}
\definecolor{mydarkblue}{rgb}{0,0.08,0.45}
\hypersetup{colorlinks=true, citecolor=mydarkblue,linkcolor=mydarkblue}

\fi

\newtheorem{theorem}{Theorem}[section]
\newtheorem{lemma}[theorem]{Lemma}
\newtheorem{definition}[theorem]{Definition}

\newtheorem{fact}[theorem]{Fact}

\newcommand{\wt}{\widetilde}
\newcommand{\ov}{\overline}

\newcommand{\R}{\mathbb{R}}

\renewcommand{\d}{\mathrm{d}}

\DeclareMathOperator*{\E}{{\mathbb{E}}}

\DeclareMathOperator{\poly}{poly}

\DeclareMathOperator{\nnz}{nnz}

\DeclareMathOperator{\rank}{rank}
\DeclareMathOperator{\diag}{diag}

\DeclareMathOperator{\ent}{ent}

\DeclareMathOperator{\reg}{reg}

\makeatletter
\newcommand*{\RN}[1]{\expandafter\@slowromancap\romannumeral #1@}
\makeatother

\usepackage{lineno}

\title{A Mathematical Abstraction for Balancing the Trade-off Between Creativity and Reality in Large Language Models}

\begin{document}

\ifdefined\isarxiv

\date{}

\title{A Mathematical Abstraction for Balancing the Trade-off Between Creativity and Reality in Large Language Models}
\author{
Ritwik Sinha\thanks{\texttt{risinha@adobe.com}. Adobe Research.}
\and 
Zhao Song\thanks{\texttt{zsong@adobe.com}. Adobe Research.}
\and
Tianyi Zhou\thanks{\texttt{t8zhou@ucsd.edu}. University of California San Diego.}
}

\else

\maketitle 
\fi

\ifdefined\isarxiv
\begin{titlepage}
  \maketitle
  \begin{abstract}

Large Language Models have become popular for their remarkable capabilities in human-oriented tasks and traditional natural language processing tasks. Its efficient functioning is attributed to the attention mechanism in the Transformer architecture, enabling it to concentrate on particular aspects of the input.

LLMs are increasingly being used in domains such as generating prose, poetry or art, which require the model to be creative (e.g. Adobe firefly). LLMs possess advanced language generation abilities that enable them to generate distinctive and captivating content. This utilization of LLMs in generating narratives shows their flexibility and potential for use in domains that extend beyond conventional natural language processing duties.

In different contexts, we may expect the LLM to generate factually correct answers, that match reality; e.g., question-answering systems or online assistants.  
In such situations, being correct is critical to LLMs being trusted in practice. 
The Bing Chatbot provides its users with the flexibility to select one of the three output modes: creative, balanced, and precise. Each mode emphasizes creativity and factual accuracy differently.

In this work, we provide a mathematical abstraction to describe creativity and reality based on certain losses. A model trained on these losses balances the trade-off between the creativity and reality of the model.

  \end{abstract}
  \thispagestyle{empty}
\end{titlepage}

{
}
\newpage

\else

\begin{abstract}

Large Language Models have become popular for their remarkable capabilities in human-oriented tasks and traditional natural language processing tasks. Its efficient functioning is attributed to the attention mechanism in the Transformer architecture, enabling it to concentrate on particular aspects of the input.

LLMs are increasingly being used in domains such as generating prose, poetry or art, which require the model to be creative (e.g. Adobe firefly). LLMs possess advanced language generation abilities that enable them to generate distinctive and captivating content. This utilization of LLMs in generating narratives shows their flexibility and potential for use in domains that extend beyond conventional natural language processing duties.

In different contexts, we may expect the LLM to generate factually correct answers, that match reality; e.g., question-answering systems or online assistants.  
In such situations, being correct is critical to LLMs being trusted in practice. 
The Bing Chatbot provides its users with the flexibility to select one of the three output modes: creative, balanced, and precise. Each mode emphasizes creativity and factual accuracy differently.

In this work, we provide a mathematical abstraction to describe creativity and reality based on certain losses. A model trained on these losses balances the trade-off between the creativity and reality of the model.
\end{abstract}

\fi

\section{Introduction}

Recently, there has been a significant development in Artificial Intelligence (AI) development. 
Large language models (LLMs) like Transformer \cite{vsp+17}, BERT \cite{dclt18}, GPT-3 \cite{bmr+20}, OPT \cite{zrg+22}  and PaLM \cite{cnd+22} have proven to be efficient methods for addressing complex natural language tasks.
OpenAI has developed ChatGPT, a chat software that relies on the 20-billion-parameter GPT-3 to produce convincing textual responses leading to compelling experiences. Additionally, the next generation GPT-4 \cite{openai23}, possesses the ability to undertake complex tasks that its predecessors were unable to perform. In various professional and academic benchmarks, GPT-4 has exhibited an outstanding level of proficiency, comparable to human performance.

The central reason behind the unprecedented success of LLMs is the \emph{attention matrix} \cite{vsp+17,rns+18,dclt18,bmr+20,as23,zhdk23,bsz23}.
An attention matrix is a table (or data structure) that shows how much each word or “token” in a text is related to the output that is being generated. This helps the model to put more attention on tokens that are related to the prediction, while reducing the effect of tokens that are not predictive. The attention matrix is learned during the training process. With the attention mechanism, each input token is checked for how much it relates to the output. This is done by giving each token a score, where this score is based on how similar the current output state and input states are. The trained models have the ability to put additional attention to the influential tokens.

More formally \cite{vsp+17}, the computation of the attention matrix starts by first multiplying the query $Q$ and key $K$ matrices together and then using a softmax function to get a resulting matrix with values ranging between $0$ and $1$. These values indicate the relative significance of each element in the input sequence. We get the attention matrix by taking the dot product of $Q$ and $K$ transpose $K^\top$, which gives us a matrix called $S$. Finally, we apply a softmax function to $S$, and this gives us our attention matrix $A$.

However, in spite of the remarkable success and popularity of language models, driven primarily by the discovery of the attention mechanism, LLMs still face several important challenges that need to be overcome. One of these stems from the fact that LLMs are likely to be ubiquitous in all types of applications from entertainment to education. On one end of the spectrum, say for writing a new script for a movie, the LLM needs to be creative. On the other end of the spectrum, e.g., for education or medical applications, the LLM needs to deliver entirely credible content which can be blindly trusted. In our work, we address this important question of how an LLM can strike the right balance between creativity and credibility. 

Creativity is essential for language models as it allows them to produce novel and diverse responses that can be useful in tasks that involve brainstorming novel directions or creating content. Demonstrating this capability is essential to establishing the intelligence of Artificial Intelligence algorithms, and for it to be an interesting experience for humans to enjoy interacting with. However, excessive creativity may result in unrealistic outputs, leading to wildly untrue statements that can destroy the credibility of the LLM. This can also limit the usefulness of the model. On the other hand, user's having confidence in the generated content is critical for LLMs to power trusted assistants to which humans can repeatedly come back for help.

Given these considerations, finding the right balance between creativity and credibility is challenging and important question for LLMs.  
\begin{center}
{\it One solution provided by Bing Chatbot is to present users with the following three options: creative, balanced, and precise. \cite{bing23} }
\end{center}
In addition to language models, our work is also relevant to other types of generative models. In particular, the idea of balancing between creativity and credibility is also relevant to text to image generation tasks. For instance, this is useful for software such as Stable Diffusion \cite{rbl+22}, DALL-E-2 from OpenAI \cite{rdn+22}, Firefly from Adobe \cite{adobe_firefly}\footnote{The Adobe Firefly is an innovative set of AI models designed to spark creativity and inspire innovation. It has the ability to generate images, text effects, and other visually stunning creations from simple text prompts.}.

Some of these software such as Bing Chatbot let users decide the balance between creativity and reality during prediction.
A natural question arises: whether that balance can be learned during training.

Our research delves into the problem of how to train the model that balances the trade-off between creativity and reality. The major idea behind our work is to describe the trade-off between creativity and reality by combining different types of losses.
\begin{definition}[Informal version of Definition \ref{def:formal_problem}]\label{def:informal_problem}
Let $\gamma \in [0,1]$. 
Let $L$ denote the loss function which can be split into two parts
\begin{align*}
   (1-\gamma) \cdot L_{\mathrm{reality}} + \gamma \cdot L_{\mathrm{creavity}} 
\end{align*}
where
\begin{itemize}
    \item The reality loss $L_{\mathrm{reality}}$ intuitively forces the model to memorize the dataset.
    \item The creativity loss $L_{\mathrm{creavity}}$ intuitively flattens the distribution of the generative model so that it has a higher chance to generate more diverse output that is further away from the data that is seen in the training data.
\end{itemize}
By tuning the ratio $\gamma $ between two losses, we can provide a trade-off to balance reality and creativity.
\end{definition}

We remark that, by tuning $\gamma \in [0,1]$ according to the size, coverage and soundness of the training dataset, the model is trained to generate responses that strike a balance between creativity and reality according to the input.
This generalization approach allows the model to adjust to a variety of situations. This results in output that is consistent with the intended level of both creativity and reality.

\subsection{Our Result}
We state our main result as the following:
\begin{theorem}[Informal]\label{thm:main:abstraction}
Let $L_{\mathrm{reality}}$ denote the $\ell_2$ regression loss, and let $L_{\mathrm{creativity}}$ denote the entropy loss. Let $\gamma$ denote the trade-off parameter between $L_{\mathrm{reality}}$ and  $L_{\mathrm{creativity}}$. For any $\gamma \in [0,1]$, there is a greedy type algorithm that starts with good initialization and runs in input sparsity time of the data, and is able to converge.

\end{theorem}

By applying our approximate Newton method, we are able to find the model that strikes a balance  between creativity and reality during the training stage.

\paragraph{Roadmap.}
We first review the previous research and literature in Section~\ref{sec:related_work}, w. In Section~\ref{sec:short_preli}, we define basic notation that is needed in the rest of the paper. In Section~\ref{sec:entropy_formulation}, we give some definitions related to entropy.
In Section~\ref{sec:technique_overview}, we provide an overview of the technique used in our paper. 
We show the Hessian of our loss function in Section~\ref{sec:short_hessian}.
In Section~\ref{sec:short_newton}, we propose an approximate Newton method that solves the optimization problem.
In Section~\ref{sec:conclusion}, we give the conclusion for our paper.


\section{Related Work}
\label{sec:related_work}

\paragraph{Large Language Models} 
Despite impressive advancements in Transformers that have led to several successful applications, our comprehension of their underlying learning mechanisms remains incomplete. Although they have demonstrated remarkable performance in structured and reasoning tasks, we still lack a comprehensive understanding of their mathematical foundations.
Prior research has proposed that the exceptional performance of Transformer-based models can be ascribed to the information that their components contain, such as the multi-head attention. Several studies \cite{tdp19, vb19, hl19, b22} have presented empirical evidence that variously suggest that these components
contain a substantial quantity of information that can assist in solving diverse probing tasks.

Various theoretical and experimental studies have been conducted to explore the full potential of Transformers. These studies have explored multiple facets of Transformers, including their ability to approximate functions \cite{ybr+20, cdw+21}, represent formal languages \cite{bag20, egz20, yppn21}, perform abstract algebraic operations \cite{zbb+22}, and achieve Turing completeness \cite{bpg20}. Some of these studies have concluded that Transformers have the potential to serve as universal approximators for operations that involve sequences and even mimic Turing machines.

In \cite{lwd+23}, contextual sparsity was found in LLM during prediction, and that sparsity pattern is found to be predictable with high accuracy. They utilized this sparsity to improve the speed of LLM inference without sacrificing performance, as supported by both theoretical analysis and empirical results. Furthermore, an alternative approach to boosting Transformer performance was introduced by \cite{dcl+21} in the form of the Pixelated Butterfly model. This model employs a straightforward fixed sparsity pattern to expedite Transformer training. Several other research studies have directed their attention towards exploring the expressiveness of the attention mechanism in Transformers \cite{dgv+18, vbc20, zkv+20, egkz21, szks21, wcm21}. 

\paragraph{Theoretical Attention Computation}

Attention computation has been the subject of various studies, examining both dynamic and static aspects. In one study \cite{bsz23}, researchers focused on dynamic attention computation and proposed an algorithm that is conditionally optimal, unless the hinted matrix vector multiplication conjecture is proven false.
They incorporated lazy update techniques into the attention computation.  Based on the Hinted MV conjecture, they show the hardness of the problem. In contrast, \cite{zhdk23,as23} concentrated on static attention computation. The authors of \cite{as23} consider the static attention computation and proposed an algorithm that established its complexity based on the exponential time hypothesis. 
\cite{zhdk23} approximate the attention in sub-quadratic time with provable spectral norm bounds.
Additionally, \cite{dms23} consider the sparsification of the attention problem and present both  deterministic and randomized algorithms. Furthermore, \cite{gsy23_dp} contributed a differentially private algorithm for computing the attention matrix. \cite{gms23} introduced a greedy descent algorithm for over-parameterized exponential regression.  \cite{lsz23} studied the approximate newton algorithm for exponential regression.

\paragraph{Newton Method and Hessian Computation}

Hessian computation is equivalent to finding the Hessian matrix, which is a square matrix of the second-order partial derivatives of the weights. The Hessian matrix describes the local curvature of a function and it is a standard task in convex optimization. Many previous works have focused on the Hessian matrix and improved many classic optimization problems, such as linear programming
\cite{lsz19,cls19,song19,b20,blss20,sy21,jswz21,dly21,b21,gs22} and sum of squares \cite{jnw22}. There are also works studied in empirical risk minimization \cite{lsz19,qszz23}, cutting plane method \cite{jlsw20,jlsz23}, semi-definite programming \cite{jkl+20,hjs+22,gs22}, and training over-parameterized neural network \cite{zmg19, cgh+19,bpsw21,szz21,hswz22,z22}.
Newton's method is an iterative optimization algorithm used to find the minimum or maximum of a function. 
In each iteration of Newton's method, the step size is determined by the inverse of the Hessian matrix. Specifically, the step size is the product of the inverse of the Hessian and the gradient vector of the function. The Hessian matrix provides information about the curvature of the function, and by taking its inverse, we can find the direction and magnitude of the step to update the estimate.

\paragraph{Creativity in Arts and Machine Learning}

To make the model outputs creative, \cite{elem17} propose a new system for generating art using Generative Adversarial Networks (GAN) to maximize deviation from established styles and minimize deviation from the art distribution. Their experiments show that the generated arts have a good balance of the trade-off between creativity and reality so that  human subjects could not distinguish art generated by the proposed system from art generated by contemporary artists.
\cite{dg07} enhance computer-generated art and make them more creatively human-like in terms of both the process and the end result. They proposed an algorithm that addresses the mentioned constraint by utilizing an automated fitness function.
In \cite{c08}, the authors propose a basic structure that can be utilized to classify and depict a program's actions, potentially enhancing the overall impression of the system's creativity.
The work \cite{rdn+22} uses contrastive models, such as CLIP \cite{openai_clip,rkh+21}, to generate images that capture both semantics and style. They demonstrate that explicitly generating image representations improves the diversity of generated images while maintaining a realistic appearance and similarity to the given caption. Building on these findings, \cite{rdn+22} proposed DALL-E-2, that employs diffusion models for the decoder.
In \cite{mda22}, the authors analyze DALL-E-2. They discuss assessing and evaluating creativity of generated art. They argue that evaluating based on output alone is insufficient and suggest considering the program's behavior and propose a simple framework to categorize and describe program behavior, enhancing the perception of creativity in the system. Adobe firefly \cite{adobe_firefly}\footnote{\url{https://www.adobe.com/sensei/generative-ai/firefly.html}} is a family of creative generative AI models coming to Adobe products, focusing on image and text generation.

\section{Preliminary}\label{sec:short_preli}

\paragraph{Basic}

For a positive integer, we use $[n]$ to denote set $\{1,2,\cdots,n\}$. We use $\E[]$ to denote expectation and $\Pr[]$ to denote probability. For two vectors $x,y$, we use $\langle x, y\rangle$ to denote the inner product. We use ${\bf 1}_n$ to denote a length-$n$ vector where all the entries are ones. For matrix $A \in \R^{n \times d}$, we use $A_{*,i}$ to denote the $i$-the column of matrix $A$ for each $i \in [d]$. For a vector $x \in \R^n$, we let $\exp(x) \in \R^n$ denote a length-$n$ vector where $\exp(x)_i= \exp(x_i)$ for each $i \in [n]$.

\paragraph{Softmax}
For convenient, we use a standard notation $\circ$ in the literature of softmax related regression \cite{lsz23,dls23,gsy23_hyper}. We use $a \circ b$ to denote a vector whose $i$-th entry is $a_i b_i$.

\paragraph{Positive Semi-definite}

For a symmetric matrix $X$, we say $X$ is positive definite ($X \succ 0$) if for each $a \in \R^d \backslash {\bf 0}_d$, we have $a^\top X a > 0$. For a symmetric matrix $X$, we say $X$ is positive semi-definite ($X \succeq 0$) if for every $a \in \R^d \backslash {\bf 0}_d$, we have $a^\top X a \geq 0$. For two symmetric matrices $X$ and $Y$, we say $X \succeq Y$, if for every $a\in \R^d$, $a^\top X a \geq a^\top Y a$.  For two PSD matrices $X$ and $Y$, we use $Y \approx_{\epsilon} X$ to denote $(1-\epsilon) X \preceq Y \preceq (1+\epsilon) X$.

\paragraph{Matrix Norms} 
For a matrix $B$, we use $\| B \|$ to denote its spectral norm. We use $\sigma_{\min}(B)$ to denote the minimum singular value of matrix $B$. We use $\sigma_{\max}(B)$ to denote the maximum singular value of matrix $B$.

\paragraph{Vector Norms}

For a vector $x$, we use $\| x \|_2$ to denote its entry-wise $\ell_2$ norm, i.e., $\| x \|_2 := ( \sum_{i=1}^n x_i^2 )^{1/2}$. For a vector $x$, we use $\| x \|_1$ to denote its entry-wise $\ell_1$ norm, i.e., $\| x \|_1 := \sum_{i=1}^n |x_i| $. We use $\| x \|_0$ to denote the number of nonzero entries in $x$. We use $\| x \|_{\infty}$ to denote $\ell_{\infty}$ norm of $x$, i.e., $\| x \|_{\infty}:= \max_{i\in [n]} |x_i|$.
It is obvious that $\| x \|_1/ \sqrt{n} \leq \| x \|_2 \leq \| x \|_1$ and $\| x \|_{\infty} \leq \| x \|_1 \leq n \| x \|_{\infty}$.

\paragraph{Input sparsity and matrix multiplication time} 

For matrix $A$, we use $\nnz(A)$ to denote the number of non-zero entries in $A$. We also refer $\nnz(A)$ time as the input sparsity time of $A$. Such notation is a well-known notation in the area of the numerical linear algebera \cite{cw13,nn13}.

We let $n^{\omega}$ denote the time of multiplying an $n \times n$ matrix with another $n \times n$ matrix.  $\omega$ in this context is referred to as the matrix multiplication exponent  \cite{w12,aw21}. Currently, we have $\omega \approx 2.373$.

\section{Mathematical Abstraction of Reality and Creativity Formulation}
\label{sec:entropy_formulation}

In Section~\ref{sec:entropy_formulation:function_definition}, we introduce the classic $\ell_2$ loss function that represents the reality part of our model.
In Section~\ref{sec:entropy_formulation:function_loss}, we introduce the entropy loss function that represents the creativity part of our model.

\subsection{A Mathematical Formulation of Reality}
\label{sec:entropy_formulation:function_definition}

We define $u(x)$ for the convenience of analysis,
\begin{definition}\label{def:u}
We define $u(x)$ as the following:
    $u(x) := \exp(Ax)$.
\end{definition}
Using standard calculus, we can show $\frac{\d u(x)}{\d x_i} = u(x) \circ A_{*,i}$ where $A_{*,i}$ denotes the $i$-th column of matrix $A$.

\begin{definition}[Normalized coefficients]
\label{def:alpha}
    We define $\alpha : \R^d \rightarrow \R$ as follows
    $
    \alpha(x) := \langle u(x), {\bf 1}_n \rangle.
    $
\end{definition}

We define function softmax $f$ as follows
\begin{definition}[Function $f$
]\label{def:f}
Given a matrix $A \in \R^{n \times d}$ and ${\bf 1}_n$ (the length-$n$ vector of ones),   
we define the prediction function $f: \R^d \rightarrow \R^n$ as follows 
$
f(x) := \langle u(x) , {\bf 1}_n \rangle^{-1} \cdot u(x) .
$
\end{definition}
For convenience, we also define the function $c(x)$ as follows. 
\begin{definition}\label{def:c}
Let $c(x) := f(x) - b$.
\end{definition}

\begin{fact}\label{fac:f}
We define $f(x)$ as Definition~\ref{def:f}. Then we have
\begin{itemize}
    \item $\langle f(x), {\bf 1}_n \rangle = 1$
    \item $\| f(x) \|_1 = 1$
    \item $\| f(x) \|_2 \leq 1$
\end{itemize}
\end{fact}
\begin{proof}
The proof is straightforward, so we omit the details of the proof here.
\end{proof}

In order to let the trained model learn to predict the correct output based on the training data. We define the $\ell_2$ loss. Recall that this represents the loss of reflecting reality with our prediction function.  
\begin{definition}\label{def:L_exp}
We define
$
    L_{\exp} := 0.5\| f(x) - b \|_2^2. 
$
\end{definition}

\subsection{A Mathematical Formulation of Creativity}\label{sec:entropy_formulation:function_loss}
We first give the definition of entropy loss that none of the  previous theoretical work \cite{dls23,lsx+23} analyzes that
\begin{definition}[Entropy]\label{def:L_ent}
We define $L_{\ent} : \R^d \rightarrow \R$, 
$
L_{\ent}(x) :=  - \langle f(x), \log f(x) \rangle.
$
\end{definition}

From the definition of entropy (Definition \ref{def:L_ent}), we know that high entropy represents a system with more randomness and uncertainty, allowing for a greater range of possibilities and thus fostering creativity.
Hence, by controlling the weight of that entropy loss, we are able to make the models' predictions have more creativity against reality. 

Next, we give a simple fact about the entropy loss.
\begin{fact}\label{fac:L_ent}
Provided that the following conditions hold
\begin{itemize}
    \item We define $f$ as Definition~\ref{def:f}.
    \item We define $L_{\ent}$  as Definition~\ref{def:L_ent}.
\end{itemize}
We have $0\leq L_{\ent}(x) \leq \log n$.
\end{fact}
 
The proof directly follow from standard calculations, for example see Fact~\ref{fac:log} in Appendix.

For convenience, we define a function $h$ as follows:
\begin{definition}[Function $h$]\label{def:h}
We define $f(x)$ as Definition \ref{def:f}. 
We define $h(x):\R^n \rightarrow \R^n$,
$
    h(x) := f(x) \circ \log f(x).
$
    
\end{definition}

\begin{fact}\label{fac:h}
We have
\begin{itemize}
    \item Part 1. $- 1 \leq h(x)_i  \leq 0$
    \item Part 2. $\| h(x) \|_{\infty} \leq 1$
    \item Part 3. $-\log n\leq \langle h(x) , {\bf 1}_n \rangle \leq 0$
    \item Part 4. $\| h(x) \|_1 \leq \log n$
    \item Part 5. $\| h(x) \|_2 \leq \log n$
\end{itemize}
\end{fact}
\begin{proof}
The proof Part 1,2 and 3 are directly follow from using Fact~\ref{fac:log} (from Appendix). Here we only provide the proofs for Part 4 and Part 5.

{\bf Proof of Part 4.}

We have
\begin{align*}
\| h(x) \|_1 =   \sum_{i=1}^n | h(x)_i | 
=  \sum_{i=1}^n - h(x)_i
=  - \langle h(x) , {\bf 1}_n \rangle 
\leq  \log n
\end{align*}
where the 1st step is using the definition of $\ell_1$ norm, the 2nd step comes from $h(x)_i \leq 0$, the 3rd step comes from the definition of inner product, and the last step comes from Part 3.

{\bf Proof of Part 5.}
We have
\begin{align*}
\| h(x) \|_2 \leq \| h(x) \|_1 \leq \log n
\end{align*}
where the 1st step is due to $\ell_2$ norm of a vector is always smaller than $\ell_1$ norm of that vector, and the 2nd step comes from Part 4.

\end{proof}

\begin{definition}[Formal version of Definition \ref{def:informal_problem}]\label{def:formal_problem}
    
We define $f(x)$ as Definition \ref{def:f}. 
We define $L_{\ent}$ as Definition \ref{def:L_ent}. Let $L_{\reg}$ denote the regularization (see Definition~\ref{def:L_reg}).
Let $b$ denote the ground truth of the data.
Let $\gamma \in [0,1]$ denote  a hyper-parameter.
The goal of our training process is to minimize the following loss function.  
\begin{align*}
    L:= (1-\gamma) \cdot L_{\exp} +  \gamma \cdot (-   L_{\ent} ) + L_{\reg}
\end{align*}
where the first term $L_{\exp}$ (which is $L_{\mathrm{reality}}$) forces the model to memorize the dataset and the second term $ -L_{\ent}$ (which is $L_{\mathrm{creativity}}$) ensures the model can output diverse responses.

\end{definition}

\section{Technique Overview}
\label{sec:technique_overview}

{\bf Hessian Computation}
Recall that the target of our paper is  to minimize the loss function $L$ (see Definition~\ref{def:informal_problem}). The Loss function $L$ can be viewed as a linear combination of reality and creativity.

In order to use the Newton method to find the solution, we first need to compute the Hessian of the loss function. In \cite{dls23}, they compute the Hessian of $L_{\exp}$. In this paper, we focus on the computation of the Hessian of $L_{\ent}$. We divide the computation into the following three steps:
\begin{itemize}
    \item Compute the gradient of $L_{\ent}$.
    \item Compute the Hessian of $\log f(x)$.
    \item Compute the Hessian of $L_{\ent}$.
\end{itemize}

In comparison with previous work \cite{lsz23,dls23,gsy23_hyper}, our approach addresses the $\log f(x)$ component in the entropy function that previous works have not handled.
To find simplify the proof, we introduce the function $h(x)$ (see Definition~\ref{def:h})
 and calculate the gradients for both $\log f(x)$ and $h(x)$.
We have 
\begin{align*}
    \frac{\d h(x)}{\d x_i} = - \langle f(x) , A_{*,i} \rangle \cdot (f(x) + h(x))+ A_{*,i}\circ (f(x) +h(x))
\end{align*}
and 
\begin{align*}
    \frac{\d \log f(x)}{ \d x_i} =  - \langle f(x) , A_{*,i} \rangle \cdot {\bf 1}_n + A_{*,i}.
\end{align*}

Subsequently, we compute Hessian matrix for $\log f(x)$, and get
\begin{align*}
        \frac{\d^2 \log f(x)}{ \d x_i^2} = (  \langle f(x), A_{*,i} \rangle^2 - \langle f(x), A_{*,i} \circ A_{*,i} \rangle ) \cdot {\bf 1}_n.
    \end{align*}
and
    \begin{align*}
        \frac{\d^2 \log f(x)}{ \d x_i \d x_j} = ( \langle f(x), A_{*,i} \rangle \cdot \langle f(x), A_{*,j} \rangle - \langle f(x), A_{*,i} \circ A_{*,j} \rangle ) \cdot {\bf 1}_n.
    \end{align*}
The detailed proof can be found in Appendix \ref{sec:gradient} and \ref{sec:hessian}.
Then, we obtain the corresponding Hessian matrix for $L_{\ent}$.

{\bf Hessian is Positive Definite}
To apply the Newton method, we also need to show that the Hessian of $L_{\ent}$ is positive  definite (We remark that in the literature of solving semi-definite programming \cite{a00,hjs+22,hjs+22_quantum}, one of the crucial step in analysis is proving the hessian is positive definite.). By decomposing the Hessian of $L_{\ent}$, we rewrite the Hessian as the following
\begin{align*}
    \frac{\d^2 L_{\ent}}{\d x_i^2} =  ~A^\top_{*,i} B(x) A_{*,i},  ~~~
    \frac{\d^2 L_{\ent} }{\d x_i x_j} = ~ A^\top_{*,i} B(x) A_{*,j},    
\end{align*}
where $B(x)$ only depends on $x$. Inspired by \cite{hjs+22}, we decompose $B(x)$ by the following two parts
$
    B(x) = B_{\mathrm{rank}}(x) + B_{\mathrm{diag}}(x),
$
where $B_{\mathrm{rank}}$ denote the low-rank parts of the decomposition of $B(x)$ and $B_{\mathrm{diag}}$ denote the diagonal parts of $B(x)$. Then, we show the bounds for both $B_{\mathrm{diag}}$ and $B_{\mathrm{rank}}$. By combining these bounds, we have 
\begin{align*}
           - 10 \log^2 n \cdot I_n \preceq  B(x) \preceq 10 \log^2 n \cdot I_n.
\end{align*}

The Hessian for $L_{\reg}$ is
$
\frac{\d^2 L_{\reg}}{ \d x^2} = A^\top W^2 A$. 
By combing the Hessian for $L_{\reg}$ and $L_{\exp}$, we have $\frac{\d^2 L }{\d x^2} \succeq l \cdot I_d$, which shows that the Hessian of the loss function is positive definite. We defer more details into Appendix~\ref{sec:hessian_psd}.

{\bf Hessian is Lipschitiz} 
Then, to show the Lipschitiz of the Hessian, we decompose $B(x)-B(y)$ as the following seven parts (see details in Appendix~\ref{sec:lipschitz_matrix}).
\begin{itemize}
    \item $G_1 = L_{\ent}(x) f(x) f(x)^\top - L_{\ent}(y) f(y) f(y)^\top$
    \item $G_2 = f(x) f(x)^\top - f(y) f(y)^\top$
    \item $G_3= f(x) h(x)^\top - f(y) h(y)^\top$
    \item $G_4 = h(x) f(x)^\top - h(y) f(y)^\top$
    \item $G_5 = \diag( L_{\ent}(x)   f(x)  ) - \diag( L_{\ent}(y)   f(y)  )$
    \item $G_6 = \diag(  f(x)  ) - \diag(   f(y)  )$
    \item $G_7 = \diag(  h(x)  ) - \diag(   h(y)  )$
\end{itemize}
where $h(x)$ is defined as $f(x) \circ \log f(x)$ (see Definition~\ref{def:h}).

By computing the Lipschitiz property for each term and combining them by using 
$
\| B(x) - B(y) \| \leq 2\| G_1 \| + \sum_{i=2}^7 \| G_i \| ,
$ (see details in Appendix~\ref{sec:hessian_lipschitz})
we have
\begin{align*}
   \| H(x) - H(y) \|  \leq  \exp(\Theta(R^2 + \log n)) \| x - y \|_2 .
\end{align*}

{\bf Approximate Newton Method}
Now, we are ready to apply the Newton method to solve the optimization problem $\min_x L(x)$. However, computing the Hessian and its inverse is expensive. Recall that the Hessian of $L$ can be written as
$
    \frac{\d^2  L}{\d x^2 }  = A^\top D A,
$
where $D =  B(x) + W^2$.
Inspired by \cite{cls19,jswz21,sy21,dsw22,syyz22}, we approximate the Hessian using the sampling matrix $\wt{D}$ based on the leverage score (see detail in Appendix \ref{sec:newton}) so that 
$
(1- \epsilon_0) A^\top D A \preceq A^\top \wt{D} A \preceq (1+\epsilon_0) A^\top D A.
$ 
Hence, we are able to reduce the running time for the expensive Hessian computation to the time of $O( (\nnz(A) + d^{\omega} )$, where $\nnz$ denote the number of non-zero entries in the matrix $A$ and  $\omega$ in this context is referred to as the matrix multiplication exponent. Currently, $w \approx 2.373$ \cite{w12,lg14,aw21}. We defer more details into Appendix~\ref{sec:newton}.

\section{Properties of Hessian}\label{sec:short_hessian}

 In Section~\ref{sec:short_hessian:pd}, we show that the Hessian is positive definite.
 In Section~\ref{sec:short_hessian:lipschitz}, we show the Lipschitz property of the Hessian.

\subsection{Hessian is Positive Definite}\label{sec:short_hessian:pd}
Recall the definition in preliminary section, for a matrix $A$, we use $\sigma_{\min}(A)$ to represent the smallest singular value of that matrix. 
We first show that the Hessian for our loss function is positive definite.
\begin{lemma}[Informal version of Lemma~\ref{lem:hessian_pd}]\label{lem:hessian_pd:informal}
Provided that the following conditions hold
\begin{itemize}
    \item Given matrix $A \in \R^{n \times d}$.
    \item Let $\gamma \in (0,1)$ denote the trade-off parameter that balances the creativity and reality. 
    \item Let $H$ be Hessian of function $L$ (see Definition~\ref{def:formal_problem}).
    \item Let $W = \diag(w) \in \R^{n \times n}$. We denote $W^2 \in \R^{n \times n}$ as the matrix that $i$-th diagonal entry is $w_{i}^2$. 
    \item Let $l > 0$ denote a scalar.
\end{itemize}
Then, we have
\begin{itemize}
    \item Part 1. If all $i \in [n]$, $w_{i}^2 \geq (1-\gamma) \cdot 10 + \gamma \cdot 10   \log^2 n+  l/\sigma_{\min}(A)^2$, then  
    \begin{align*}
    H \succeq l \cdot I_d
    \end{align*}
    \item Part 2. If all $i \in [n]$, $w_{i}^2 \geq (1-\gamma) \cdot 200  +  \gamma \cdot 200 \log^2 n + l/\sigma_{\min}(A)^2$, then  
    \begin{align*}
        W^2 \approx_{1/10} B(x) +W^2. 
    \end{align*}
\end{itemize}
\end{lemma}
Next, we will move to discuss the Lipschitz property of Hessian matrix.

\subsection{Hessian is Lipschitz}\label{sec:short_hessian:lipschitz}
Then, we show the Lipschitz property of the Hessian.
\begin{lemma}[Informal version of Lemma~\ref{lem:hessian_lipschitz}]\label{lem:hessian_lipschitz:informal}
Provided the two conditions below:
\begin{itemize}
    \item Let $\max\{ \| x \|_2, \|y \|_2, \| A \| \} \leq R$
    \item Let $H$ be Hessian of loss function $L$ (see Definition~\ref{def:formal_problem})
\end{itemize}
Then, we have
    $
    \| H(x) - H(y) \| \leq n^2 \exp(40R^2) \| x - y \|_2
    $.
\end{lemma}

\section{Approximate Newton Method}\label{sec:short_newton}

In this section, we propose our approximate Newton method that solves the optimization problem defined in Definition \ref{def:formal_problem}.

We first provide the following conditions that are required for the approximate Newton method. 
\begin{definition}[Properties of Loss Function 
]\label{def:assumptions}
Given a function $L : \R^d \rightarrow \R$, we call function $L$ $(l,M)$-good if it satisfies the following conditions:
\begin{itemize}
    \item Property 1. 
    Let $l$ denote a positive scalar. If there is a vector $x^* \in \R^d$ satisfying that:
    \begin{itemize}
        \item $\nabla L(x^*) = {\bf 0}_d$.
        \item $\nabla^2 L(x^*) \succeq l \cdot I_d$.
    \end{itemize}
    \item Property 2. 
    Suppose $M>0$ satisfying that
    $
        \| \nabla^2 L(y) - \nabla^2 L(x) \| \leq M \cdot \| y - x \|_2 
        $
    \item Property 3. 
    Let $x_0$ denote the initialization point. If $\mathrm{res}_0:=\| x_0 -x^*\|_2$ satisfies
   $
        \mathrm{res}_0  \cdot M  \leq 0.1 l.
    $   
\end{itemize}
\end{definition}
Property 1 is describing the local minimum. Property 2 is indicating the Lipschitz property of Hessian, and property 3 guarantees that we start a from reasonably good starting point.

By using the sampling matrix based on the leverage score (see Lemma \ref{lem:subsample} for detail), we have the following approximate Newton update (see Definition \ref{def:update_x_t} for detail):
\begin{align*}
    x_{t+1} = x_t  -  \mathrm{Approximate Hessian}(x_t)^{-1} \cdot  \mathrm{gradient}(x_t)  .
\end{align*}

Recall from the preliminary section.  We use ${\bf 0}_n$ to represent a length-$n$ vector where all the values are zeros. We use $\omega \approx 2.373$ to represent the exponent of matrix multiplication. Suppose that $\sigma_{\min}(A)$ is the smallest singular value of matrix $A$. 
Next, we introduce our main result using the approximate Newton method. 
\begin{theorem}[Formal version of Theorem~\ref{thm:main:abstraction}]\label{thm:main_informal}
Given matrix $A \in \R^{n \times d}$, $b \in \R^n$, and $w \in \R^n$. 
\begin{itemize}
    \item We define $f(x):=\langle \exp(Ax), \mathbf{1}_n \rangle^{-1} \exp(Ax)$. We define $L_{\exp}$ as Definition~\ref{def:L_exp}. We define $L_{\ent}$ as Definition~\ref{def:L_ent} and $L_{\reg}$ as Definition~\ref{def:L_reg}.
    \item Suppose that $R \geq 10$. Let $l > 0$. Let $\gamma \in [0,1]$. Let $M = \exp(\Theta( R^2 + \log n))$.
    \item We use $x^*$ to denote the optimal solution of  
    $
    \min_{x \in \R^d} (1-\gamma) \cdot L_{\exp} - \gamma \cdot L_{\ent} + L_{\reg}
    $
    that
     $g(x^*) = {\bf 0}_d$, $\| x^* \|_2 \leq R$.
        
    \item Assume that $\| A \| \leq R$. Here $\| A \|$ denotes the spectral norm of matrix $A$.
    
    \item Suppose that $b \geq {\bf 0}_{n}$ and $\| b \|_1 \leq 1$. (Here $b \geq {\bf 0}_n$ denotes $b_i \geq 0$ for all $i\in[n]$) 

    \item Assume that  $w_{i}^2 \geq (1-\gamma) \cdot 100  +  \gamma \cdot 100 \log^2 n + l/\sigma_{\min}(A)^2$ for all $i \in [n]$.
    \item Let $x_0$ denote an starting/initial point such that $M \| x_0 - x^* \|_2 \leq 0.1 l$.
    \item Let $\delta \in (0,0.1)$ and  $\epsilon \in (0,0.1)$ be our failure probability and accuracy parameter respectively.
\end{itemize}
There is a randomized algorithm (Algorithm~\ref{alg:main}) that
\begin{itemize}
\item  runs $\log(\| x_0 - x^* \|_2/ \epsilon)$ iterations 
\item spend  
$
O( (\nnz(A) + d^{\omega} ) \cdot \poly(\log(n/\delta)) 
$
time per iteration,
\item and finally outputs a vector $\wt{x} \in \R^d$ such that
$
\Pr[ \| \wt{x} - x^* \|_2 \leq \epsilon ] \geq 1-\delta.
$
\end{itemize}
\end{theorem}

\section{Conclusion}\label{sec:conclusion}

Large Language Models (LLMs) have demonstrated their impressive capabilities and strong performance in human-oriented tasks and natural language processing. The Transformer architecture's attention mechanism allows LLMs to selectively focus on specific input elements. LLMs have been utilized effectively in creative language generation, which demonstrates their potential in non-traditional fields. The Bing Chatbot produces output modes that balance creativity and factual accuracy. This work proposes a mathematical abstraction to balance creativity and reality in LLMs and proposed an algorithm that solves the optimization problem using the Newton method.
It is crucial to note that our conclusions were drawn from simplified Transformer models and simple regression problems. Further research is necessary to gain a comprehensive understanding of the balance between creativity and reality in larger models. Our work, being theoretical in nature, is not expected to have any negative societal impact.

\ifdefined\isarxiv

\else
\bibliography{ref}
\bibliographystyle{plain}

\fi

\newpage
\onecolumn
\appendix

\section*{Appendix}

{\bf Roadmap.}

In Section \ref{sec:preliminary}, we introduce the necessary background and tools.
In Section \ref{sec:gradient}, we derive the gradient of the entropy loss function.
In Section \ref{sec:hessian}, we compute the Hessian of the entropy loss function.
In Section \ref{sec:hessian_psd}, we show that the Hessian of the entropy loss function is PSD.
In Section \ref{sec:hessian_lipschitz},  we show that the Hessian of the entropy loss function is Lipschlitz.
In Section \ref{sec:lipschitz_scalor}, we compute the Lipschitz property for $L_{\ent}$ and give an upper bound on $\|\log(1/f(x))\|_\infty$.
In Section \ref{sec:lipschitiz_vector}, we show the Lipschitz property for some vector functions.
In Section \ref{sec:lipschitz_matrix}, we show the Lipschitz property for some matrix functions.
In Section \ref{sec:newton}, we provide the definition of $(l, M )$-good Loss function and show the approximation of Hessian and the update rule.
In Section \ref{sec:main_result}, we introduce our main results.

\section{Preliminary}
\label{sec:preliminary}

\begin{fact}\label{fac:diag_circ}
For any vectors $a, b \in \R^n$, we have
\begin{itemize}
    \item $a\circ a^{-1} = {\bf 1}_n$
    \item $a \circ a^{-1} \circ b = b$
\end{itemize}
\end{fact}

\begin{fact}\label{fac:log}
We have
\begin{itemize}
    \item Part 1. For any $z \in (0,1)$, we have $z \log (1/z) \in (0,1)$
    \item Part 2. For any $z \in (0,1)$ we have $z \log z \in (-1,0)$
\end{itemize}
For a vector $x \in \R_{> 0}^n$ with $\sum_{i=1}^n x_i = 1$, we have
\begin{itemize}
    \item Part 3. $\sum_{i=1}^n x_i \log(1/x_i) \leq \log n$
\end{itemize}
We have 
\begin{itemize}
    \item Part 4. Taylor series of $\log(1+z)$ at $z=0$ is $\sum_{i=1}^{\infty} (-1)^{i+1} \frac{1}{i} x^i$
    \item Part 5. For $|z| \leq 0.1 $, we have $| \log(1+z) | \leq |z|$
    \item Part 6. For $| (x-y)/x | \leq 0.1$, we have $|\log(y/x)| = \log(1 + \frac{y-x}{x}) \leq | \frac{y-x}{x} |$
\end{itemize}
\end{fact}
\begin{proof}

{\bf Proof of Part 3.}

The function chooses the largest possible value when all $x_i = 1/n$.
\end{proof}

\begin{fact}\label{fac:basic_vector_norm}
We have
\begin{itemize}
    \item $\| a \circ b \|_2 \leq \| a \|_2 \cdot \| b \|_{\infty}$
    \item $\langle a, b \rangle \leq \| a \|_2 \cdot \| b \|_2$
\end{itemize}

\end{fact}

\begin{fact}\label{fac:basic_matrix_norm}
We have
\begin{itemize}
    \item For two vectors $a,b$ we have $\| a b^\top \| \leq \| a \|_2 \cdot \| b \|_2$.
    \item For a scalar $\alpha$ and a matrix $B$, we have $| \alpha B \| \leq |\alpha| \cdot \| B \|$
\end{itemize}
\end{fact}

\begin{fact}
For any vector $u,v \in \R^n, \|u-v\|_{\infty} \leq 0.01$, we have
\begin{itemize}
    \item $\|\exp (u)-\exp (v)\|_2 \leq\|\exp (u)\|_2 \cdot 2\|u-v\|_{\infty}$
\end{itemize}

\end{fact}
\begin{proof}
We have
\begin{align*}
\|\exp (u)-\exp (v)\|_2 & = \|\exp (u) \circ (\mathbf{1}_n-\exp (v-u))\|_2 \\
& \leq\|\exp (u)\|_2 \cdot \|\mathbf{1}_n-\exp (v-u)\|_{\infty} \\
& \leq\|\exp (u)\|_2 \cdot 2\|u-v\|_{\infty}
\end{align*}
where the 1st step is using definition of $\circ$ operation and $\exp ()$, the 2nd step is due to $\|u \circ v\|_2 \leq\|u\|_{\infty} \cdot\|v\|_2$, the 3rd step is due to $|\exp (x)-1| \leq 2 x$ for all $x \in (0,0.1)$.

\end{proof}

\subsection{Regularization}\label{sec:preli:reg}

Adding regularization is a common trick in many machine learning tasks \cite{kmn+16,hhs17,gdg+17,shn+18,dssw18,ll19,achl19,cb20,jmgl21,wl21,jrg22}.
\cite{kmn+16} using data augmentation to add regularization to the training process. They also combine ADAM and the $\ell_2$ norm regularization.
\cite{hhs17,gdg+17} introduce how increasing the batch size can help regularize the training process. 
\cite{shn+18} aid in understanding implicit regularization in  complex models.
\cite{dssw18} apply the regularization to solve the Kronecker product regression.
\cite{ll19} study the implicit regularization of the gradient descent algorithm in homogeneous neural networks
\cite{achl19} study the implicit regularization of gradient descent over deep linear neural networks for matrix completion and sensing.
Regularization has been applied in numerous machine learning scenarios. In the field of binary classification, \cite{cb20} employs algorithmic regularization techniques. Similarly, in contrastive learning, \cite{wl21} utilize algorithmic regularization methods. Additionally, in the context of convolutional neural networks, \cite{jrg22} apply algorithmic regularization approaches.

\begin{definition}\label{def:L_reg}
Given matrix $A \in \R^{n \times d}$.
For a given vector $w \in \R^n$, let $W = \diag(w)$. 
We define $L_{\reg} : \R^d \rightarrow \R$ as follows
\begin{align*}
L_{\reg}(x):= \frac{1}{2} \| W A x\|_2^2
\end{align*}
\end{definition}

\begin{lemma}[Folklore, see \cite{lsz23} as an example]\label{lem:L_reg_gradient_hessian}
For a given vector $w \in \R^n$, let $W = \diag(w)$. Let $L_{\reg} : \R^d \rightarrow \R$ be defined as Definition~\ref{def:L_reg}.

Then, we have
\begin{itemize}
\item The gradient is
\begin{align*}
\frac{\d L_{\reg}}{ \d x} = A^\top W^2 Ax
\end{align*}
\item The Hessian is
\begin{align*}
\frac{\d^2 L_{\reg}}{ \d x^2} = A^\top W^2 A
\end{align*}
\end{itemize}
\end{lemma}
\section{Gradient}\label{sec:gradient}
In Section~\ref{sec:gradient:u}, we calculate the gradient of the function $u(x)$.
In Section~\ref{sec:gradient:f}, we compute the gradient of $f(x)$. 
Moving on to Section~\ref{sec:gradient:log_f}, we analyze the gradient of $\log f(x)$ by using the gradient of $f(x)$. 
Next, in Section~\ref{sec:gradient:h}, we compute the gradient of $h(x)$.  
In Section~\ref{sec:gradient:ent}, we compute the gradient of $L_{\ent}$. 
Building on the gradient computation in the previous sections we compute the gradient of $L$  in Section~\ref{sec:gradient_loss_L}.

\subsection{Gradient of \texorpdfstring{$u(x)$}{}}\label{sec:gradient:u}

The goal of this section is to present Lemma~\ref{lem:gradient_u}. 
\begin{lemma}[]\label{lem:gradient_u}
Provided that the following conditions are met
\begin{itemize}
\item Let $u(x)$ be defined as Definition~\ref{def:u}.
\end{itemize}
Then for every $i \in [d]$, we have
\begin{itemize}
    \item 
    \begin{align*}
        \frac{\d u(x)}{ \d x_i} = u(x) \circ A_{*,i}
    \end{align*}
\end{itemize}
\end{lemma}
\begin{proof}
The proof follows from Lemma 5.6  in \cite{dls23}.
\end{proof}

\subsection{Gradient of \texorpdfstring{$f(x)$}{}}\label{sec:gradient:f}
The goal of this section is to derive the gradient of $f(x)$ (Lemma \ref{lem:gradient_f}).
\begin{lemma}[Basic gradient computation]\label{lem:gradient_f}
If the following conditions hold
\begin{itemize}
    \item Given matrix $A \in \R^{n \times d}$.
    \item Given vector $b \in \R^n$.
    \item We define $\alpha : \R^d \rightarrow \R$ as Definition~\ref{def:alpha}.
    \item We define $f : \R^d \rightarrow \R^n$ as Definition~\ref{def:f}.
    \item We define $L_{\ent} : \R^d \rightarrow \R$ as Definition~\ref{def:L_ent}.
\end{itemize}
 For every $i \in [d]$, we have
 \begin{itemize}
    \item Part 1. 
    \begin{align*}
        \frac{\d f (x) }{\d x_i}  =  & ~ -  \langle f(x), A_{*,i}\rangle \cdot f(x) + f(x) \circ A_{*,i}
    \end{align*} 
    \item Part 2. 
    \begin{align*}
        \langle \frac{\d f(x) }{\d x_i} , A_{*,i} \rangle = - \langle f(x), A_{*,i} \rangle^2 + \langle f(x), A_{*,i} \circ A_{*,i} \rangle
    \end{align*}
    \item Part 3. 
    \begin{align*}
        \langle \frac{\d f(x) }{\d x_i} , A_{*,j} \rangle = - \langle f(x), A_{*,i} \rangle \cdot \langle f(x), A_{*,j} \rangle + \langle f(x), A_{*,i} \circ A_{*,j} \rangle
    \end{align*}
    \begin{itemize}
        \item This is used at {\bf Part 3} of Lemma \ref{lem:gradient_ent}
        \item This is used at {\bf Part 1} of Lemma \ref{lem:hessian_log_f}
        \item This is used at {\bf Part 2} of Lemma \ref{lem:hessian_log_f}
    \end{itemize}
    \item Part 4. 
    \begin{align*}
        \langle \frac{\d f(x)}{ \d x_i} , \log f(x) \rangle =   \langle f(x), A_{*,i}\rangle \cdot L_{\ent}(x) +  \langle  h(x), A_{*,i}\rangle 
    \end{align*}
    \begin{itemize}
        \item This is used at {\bf Part 1} of  Lemma~\ref{lem:gradient_ent}
    \end{itemize}
    \item Part 5. 
    \begin{align*}
        ( \frac{\d f(x)}{ \d x_i} ) \circ \log f(x) =  -  \langle f(x), A_{*,i}\rangle \cdot h(x) + h(x) \circ A_{*,i}
    \end{align*}
    \begin{itemize}
        \item This is used at {\bf Part 1} of  Lemma~\ref{lem:gradient_h}
    \end{itemize}
\end{itemize}
\end{lemma}
\begin{proof}

{\bf Proof of Part 1, 2, 3}
The proof directly follows from Lemma 5.6 in \cite{dls23}.

{\bf Proof of Part 4.}

We have 
\begin{align*}
    \langle \frac{\d f(x)}{ \d x_i} , \log f(x) \rangle  = & ~ \langle -\langle f(x), A_{*,i}\rangle \cdot f(x) + f(x) \circ A_{*,i} , \log f(x) \rangle\\
    = & ~ -\langle \langle f(x), A_{*,i}\rangle \cdot f(x), \log f(x) \rangle + \langle f(x) \circ A_{*,i} , \log f(x) \rangle\\
    = & ~-\langle f(x), A_{*,i}\rangle \cdot  \langle  f(x), \log f(x) \rangle + \langle f(x) \circ A_{*,i} , \log f(x) \rangle\\
    = & ~  \langle f(x), A_{*,i}\rangle \cdot L_{\ent} +  \langle  h(x), A_{*,i}\rangle  
\end{align*}
where the 1st step comes from Part 1, the 2nd step follows from simple algebra, the 3rd step comes from simple algebra and the last step follows from Definition \ref{def:L_ent} and Definition \ref{def:h}.

{\bf Proof of Part 5.}

\begin{align*}
      ( \frac{\d f(x)}{ \d x_i} ) \circ \log f(x)  = & ~ (-  \langle f(x), A_{*,i}\rangle \cdot f(x) + f(x) \circ A_{*,i}) \circ \log f(x)\\
      = & ~ -  \langle f(x), A_{*,i} \rangle \cdot f(x) \circ \log f(x) + f(x) \circ A_{*,i} \circ \log f(x) \\
      = & ~ -  \langle f(x), A_{*,i}\rangle \cdot h(x) + h(x) \circ A_{*,i}
\end{align*}
where the 1st  step comes from Part 1, the 2nd step follows from simple algebra and the 3rd step comes from Definition \ref{def:h}.

\end{proof}

\subsection{Gradient for \texorpdfstring{$\log f(x)$}{}}\label{sec:gradient:log_f}

Building on the gradient of $f(x)$, in this section, we calculate the gradient of $\log f(x)$.
\begin{lemma}\label{lem:gradient_log_f}
Provided that the following condition holds
\begin{itemize}
    \item We define $f$ as Definition~\ref{def:f}.
\end{itemize}
Then, we have
\begin{itemize}
\item Part 1.
\begin{align*}
    \frac{\d \log f(x)}{ \d x_i} =  - \langle f(x) , A_{*,i} \rangle \cdot {\bf 1}_n + A_{*,i}
\end{align*}
    \begin{itemize}
        \item This is used at {\bf Part 1} of  Lemma~\ref{lem:hessian_log_f}
        \item This is used at {\bf Part 2} of  Lemma~\ref{lem:hessian_log_f}
    \end{itemize}
\item Part 2.
\begin{align*}
 \langle \frac{\d \log f(x)}{ \d x_i} , b \rangle = \langle A_{*,i} , b \rangle- \langle f(x), A_{*,i} \rangle \cdot \langle b, {\bf 1}_n \rangle 
\end{align*}
\item Part 3.
\begin{align*}
 \langle \frac{\d \log f(x)}{ \d x_i} , f(x) \rangle  = 0
\end{align*}
\item Part 4. 
\begin{align*}
    (\frac{\d \log f(x) }{\d x_i} ) \circ f(x) = - \langle f(x) , A_{*,i} \rangle \cdot f(x) + A_{*,i}\circ f(x) 
\end{align*}
\end{itemize}
\end{lemma}
\begin{proof}

{\bf Proof of Part 1.}

For every $j \in [n]$, we have
\begin{align*}
\frac{\d \log f(x)_j}{ \d x_i} = f(x)_j^{-1} \frac{\d f(x)_j}{ \d x_i}
\end{align*}
Then we group the $n$ coordinates, we get
\begin{align*}
\frac{\d \log f(x)}{ \d x_i} 
= & ~ f(x)^{-1} \circ \frac{\d f(x)}{ \d x_i} \\
= & ~ f(x)^{-1} \circ ( -  \langle f(x), A_{*,i}\rangle f(x) + f(x) \circ A_{*,i} ) \\
= & ~ -\langle f(x), A_{*,i} \rangle f(x)^{-1} \circ f(x) + f(x)^{-1} \circ f(x) \circ A_{*,i} \\
= & ~ - \langle f(x) , A_{*,i} \rangle \cdot {\bf 1}_n + A_{*,i}
\end{align*}

{\bf Proof of Part 2.}
We have 
\begin{align*}
     \langle \frac{\d \log f(x)}{ \d x_i} , b \rangle = & ~ \langle - \langle f(x) , A_{*,i} \rangle \cdot {\bf 1}_n + A_{*,i}, b \rangle \\
     = & ~\langle A_{*,i} , b \rangle- \langle f(x), A_{*,i} \rangle \cdot \langle b, {\bf 1}_n \rangle ,
\end{align*}
where the 1st step follows from Part 1 and the reason for the last step is simple algebra.

{\bf Proof of Part 3.}
We have
\begin{align*}
       \langle \frac{\d }{\d x_i} (\log f(x)),  f(x)\rangle = & ~  \langle - \langle f(x) , A_{*,i} \rangle \cdot {\bf 1}_n + A_{*,i},  f(x)\rangle\\
        = & ~ -\langle f(x) , A_{*,i} \rangle \cdot  \langle    {\bf 1}_n,  f(x) \rangle + \langle A_{*,i}, f(x)\rangle\\
        = & ~ -\langle f(x) , A_{*,i} \rangle+ \langle A_{*,i}, f(x)\rangle   \\
        = & ~  0
\end{align*}
where third step follows from $\langle {\bf 1}_n, f(x) \rangle =1$ (Fact~\ref{fac:f}).

{\bf Proof of Part 4.}

We have
\begin{align*}
      (\frac{\d \log f(x) }{\d x_i} ) \circ f(x) = & ~ (- \langle f(x) , A_{*,i} \rangle \cdot {\bf 1}_n + A_{*,i}) \circ f(x)\\
      = & ~ - \langle f(x) , A_{*,i} \rangle \cdot f(x) + A_{*,i}\circ f(x) 
\end{align*}
where the first step comes from Part 1 and the second step follows from simple algebra.
\end{proof}

\subsection{Gradient for \texorpdfstring{$h(x)$}{}}\label{sec:gradient:h}

The goal of this section is to give Lemma \ref{lem:gradient_h}.
\begin{lemma}\label{lem:gradient_h}
If the following conditions hold
\begin{itemize}
    \item We define function $f : \R^d \rightarrow \R^n$ as Definition~\ref{def:f}.
    \item We define function $h : \R^d \rightarrow \R^n$ as Definition \ref{def:h}.
\end{itemize}    
We have
    \begin{itemize}
        \item Part 1.
            \begin{align*}
        \frac{\d h(x)}{\d x_i} = - \langle f(x) , A_{*,i} \rangle \cdot (f(x) + h(x))+ A_{*,i}\circ (f(x) +h(x))
    \end{align*}
\item Part 2.  
\begin{align*}
    \frac{\d}{\d x_i} \langle h(x), A_{*,i}\rangle =  - \langle f(x) , A_{*,i} \rangle \cdot \langle f(x) + h(x), A_{*,i} \rangle + \langle f(x) +h(x),  A_{*,i} \circ A_{*,i} \rangle
\end{align*}

\item Part 3. 
\begin{align*}
    \frac{\d}{\d x_i} \langle h(x), A_{*,j}\rangle = - \langle f(x) , A_{*,i} \rangle \cdot \langle f(x) + h(x), A_{*,j} \rangle + \langle f(x) +h(x),  A_{*,i} \circ A_{*,j} \rangle 
\end{align*}
    \end{itemize}
\end{lemma}
\begin{proof}

{\bf Proof of Part 1.}

For every $i \in [n]$, we have
    \begin{align*}
        \frac{\d h(x)}{\d x_i} = & ~ \frac{\d }{\d x_i}( \log f(x) \circ f(x)) \\
        = & ~ ( \frac{\d }{\d x_i}  \log f(x) ) \circ f(x) +  \log f(x) \circ ( \frac{\d }{\d x_i} f(x) ) \\
        = & ~  - \langle f(x) , A_{*,i} \rangle \cdot f(x) + A_{*,i}\circ f(x)   -  \langle f(x), A_{*,i}\rangle \cdot h(x) + h(x) \circ A_{*,i}\\
        = & ~  - \langle f(x) , A_{*,i} \rangle \cdot (f(x) + h(x))+ A_{*,i}\circ (f(x) +h(x))
    \end{align*}
    where the first step comes from Definition \ref{def:h}, the second step follows from chain rule, the third step comes from Part 5 of Lemma \ref{lem:gradient_f} and Part 4 of Lemma \ref{lem:gradient_log_f}, and the reason for last step is simple algebra.

{\bf Proof of Part 2.}
It trivially comes from Part 1.
\begin{align*}
    \frac{\d}{\d x_i} \langle h(x), A_{*,i}\rangle = & ~ - \langle f(x) , A_{*,i} \rangle \cdot \langle f(x) + h(x), A_{*,i} \rangle + \langle f(x) +h(x),  A_{*,i} \circ A_{*,i} \rangle 
\end{align*}

{\bf Proof of Part 3.}
Following the proof of Part 1, we have
\begin{align*}
    \frac{\d}{\d x_i} \langle h(x), A_{*,j}\rangle = & ~ - \langle f(x) , A_{*,i} \rangle \cdot \langle f(x) + h(x), A_{*,j} \rangle + \langle f(x) +h(x),  A_{*,i} \circ A_{*,j} \rangle 
\end{align*}
Thus, we complete the proof.
\end{proof}

\subsection{Gradient for \texorpdfstring{$L_{\ent}(x)$}{}}\label{sec:gradient:ent}

The goal of this section is to present Lemma \ref{lem:gradient_ent}.
\begin{lemma}\label{lem:gradient_ent}
If the following condition holds
\begin{itemize}
    \item  We define  $f : \R^d \rightarrow \R^n$  as Definition~\ref{def:f}.
    \item  We define  $L_{\ent} : \R^d \rightarrow \R$ as Definition \ref{def:L_ent}.
    \item  We define  $h : \R^d \rightarrow \R^n$ as Definition~\ref{def:h}
    \item  We define  $\alpha$  as Definition \ref{def:alpha}.
\end{itemize}
We have

\begin{itemize}
    \item Part 1.
    \begin{align*}
        \frac{\d L_{\ent}(x)}{\d x_i}  =-\langle f(x), A_{*,i}\rangle L_{\ent}(x) - \langle h(x), A_{*,i}\rangle 
    \end{align*}
    \item Part 2.
    \begin{align*}
        \frac{\d}{\d x_i} ( \langle f(x), A_{*,i}\rangle \cdot L_{\ent}(x)) = & ~  -2\langle f(x), A_{*,i}\rangle^2 \cdot L_{\ent}(x)+ \langle f(x), A_{*,i} \circ A_{*,i} \rangle \cdot L_{\ent}(x) \\
    & ~ -  \langle f(x), A_{*,i}\rangle \cdot \langle h(x), A_{*,i}\rangle 
    \end{align*}
    \item  Part 3. 
    \begin{align*}
        \frac{\d}{\d x_i} ( \langle f(x), A_{*,j}\rangle \cdot L_{\ent}(x)) = & ~  -2 \langle f(x), A_{*,j}\rangle \cdot \langle f(x), A_{*,i}\rangle L_{\ent} (x)+ \langle f(x), A_{*,i} \circ A_{*,j} \rangle \cdot L_{\ent}(x) \\
    & ~ - \langle f(x), A_{*,j}\rangle \cdot \langle h(x), A_{*,i}\rangle 
    \end{align*}
\end{itemize}
\end{lemma}
\begin{proof}

{\bf Proof of Part 1.}
We have
    \begin{align*}
        \frac{\d L_{\ent}(x)}{\d x_i} = & ~ - \frac{\d }{\d x_i}\langle \log f(x),  f(x)\rangle \\
        = & ~ - \langle \frac{\d }{\d x_i} (\log f(x)),  f(x)\rangle - \langle  \log f(x),  \frac{\d }{\d x_i} f(x)\rangle \\ 
        = & ~ 0 - \langle  \log f(x),  \frac{\d }{\d x_i} f(x)\rangle \\
        = & ~ -\langle f(x), A_{*,i}\rangle \cdot L_{\ent}(x) - \langle h(x), A_{*,i}\rangle 
    \end{align*}
    where the 1st step is due to Definition \ref{def:L_ent}, the 2nd step follows from the chain rule and the 3rd step is using {\bf Part 3} of Lemma \ref{lem:gradient_log_f}, the 4th step is using {\bf Part 4} of Lemma \ref{lem:gradient_f}.

{\bf Proof of Part 2.}
We have 
\begin{align*}
    \frac{\d}{\d x_i} ( \langle f(x), A_{*,i}\rangle \cdot L_{\ent}(x)) 
    = & ~  ( \frac{\d}{\d x_i}  \langle f(x), A_{*,i}\rangle ) \cdot L_{\ent}(x) +     \langle f(x), A_{*,i}\rangle \cdot \frac{\d}{\d x_i} L_{\ent}(x) \\
    = & ~ (- \langle f(x), A_{*,i} \rangle^2 + \langle f(x), A_{*,i} \circ A_{*,i} \rangle) \cdot L_{\ent}(x)\\
    & ~ + \langle f(x), A_{*,i}\rangle \cdot (- \langle f(x), A_{*,i}\rangle L_{\ent} (x)- \langle h(x), A_{*,i}\rangle ) \\
    = & ~ - \langle f(x), A_{*,i} \rangle^2 \cdot L_{\ent}(x) + \langle f(x), A_{*,i} \circ A_{*,i} \rangle \cdot L_{\ent}(x) \\
    & ~ -   \langle f(x), A_{*,i}\rangle^2 \cdot L_{\ent}(x) - \langle f(x), A_{*,i}\rangle \cdot \langle h(x), A_{*,i}\rangle  \\
    = & ~-2\langle f(x), A_{*,i}\rangle^2 \cdot L_{\ent}(x)+ \langle f(x), A_{*,i} \circ A_{*,i} \rangle \cdot L_{\ent}(x) \\
    & ~ -  \langle f(x), A_{*,i}\rangle \cdot \langle h(x), A_{*,i}\rangle  
\end{align*}
where the 1st step is using the chain rule, the 2nd step is due to Part 1 above and Part 2 in Lemma \ref{lem:gradient_f}, the 3rd step is using simple algebra and the reason for last step is simple algebra.

{\bf Proof of Part 3.}
We have 
\begin{align*}
    \frac{\d}{\d x_i} ( \langle f(x), A_{*,j}\rangle \cdot L_{\ent}(x)) 
    = & ~   \frac{\d}{\d x_i}  \langle f(x), A_{*,j}\rangle  \cdot L_{\ent}(x) +     \langle f(x), A_{*,j}\rangle \cdot \frac{\d}{\d x_i} L_{\ent}(x) \\
    = & ~ (- \langle f(x), A_{*,i} \rangle \cdot \langle f(x), A_{*,j} \rangle + \langle f(x), A_{*,i} \circ A_{*,j} \rangle) \cdot L_{\ent}(x)\\
    & ~ - \langle f(x), A_{*,j}\rangle \cdot (\langle f(x), A_{*,i}\rangle L_{\ent} (x)- \langle h(x), A_{*,i}\rangle ) \\
    = & ~  -2 \langle f(x), A_{*,j}\rangle \cdot \langle f(x), A_{*,i}\rangle L_{\ent} (x)+ \langle f(x), A_{*,i} \circ A_{*,j} \rangle \cdot L_{\ent}(x) \\
    & ~ - \langle f(x), A_{*,j}\rangle \cdot \langle h(x), A_{*,i}\rangle  
\end{align*}
where the 1st step is using the chain rule, the 2nd step is due to Part 1 above and Part 3 in Lemma \ref{lem:gradient_f} and the reason for the last step is simple algebra.

\end{proof}

\subsection{Gradient for Loss \texorpdfstring{$L$}{}}\label{sec:gradient_loss_L}

Building on the gradient calculation of the above sections, we are now ready to derive the gradient of our loss function $L(x)$.
\begin{lemma}\label{lem:gradient_loss_L}
If the following conditions holds
\begin{itemize}
    \item  We define  $L_{\ent}$ as Definition~\ref{def:L_ent}.
    \item  We define  $L_{\exp}$ as Definition~\ref{def:L_exp}.
    \item   We define  $L_{\reg}$ as Definition~\ref{def:L_reg}.
    \item  We define  $h : \R^d \rightarrow \R^n$ as Definition~\ref{def:h}
    \item   We define  $L$ as $L(x) = (1-\gamma) \cdot L_{\exp}(x) + \gamma \cdot (- L_{\ent}(x) ) + L_{\reg}(x)$. 
\end{itemize}
Then we have
\begin{itemize}
\item Part 1.
\begin{align*}
    \frac{\d L}{ \d x_i} = & ~ (1-\gamma) (A_{*,i}^\top \cdot (f(x) (f(x) - b)^\top f(x) + \diag(f(x))(f(x)-b)) \\
    & ~ + \gamma ( \langle f(x), A_{*,i}\rangle L_{\ent}(x) + \langle h(x), A_{*,i}\rangle) \\
    & ~ + A_{*,i}^\top W^2 Ax
\end{align*}
\item Part 2.
\begin{align*}
\frac{\d L}{ \d x} = & ~ \underbrace{ (1-\gamma) }_{ \mathrm{scalar} } \cdot ( \underbrace{ A^\top }_{d \times n} \cdot \underbrace{ (f(x) (f(x) - b)^\top f(x) + \diag(f(x))(f(x)-b) }_{n \times 1} ) \\
& ~ + \underbrace{ \gamma }_{ \mathrm{scalar} } \cdot \underbrace{ (A^\top f(x) L_{\ent}(x) + A^\top h(x) ) }_{d \times 1} \\
& ~ + \underbrace{ A^\top W^2 A  x }_{d \times 1}
\end{align*}
\end{itemize}
\end{lemma}
\begin{proof}

{\bf Proof of Part 1.}
We can show
\begin{align*}
    \frac{\d L}{ \d x_i} = & ~ \frac{\d}{ \d x_i}  (1-\gamma) \cdot L_{\exp}(x) - \gamma \cdot L_{\ent}(x) + L_{\reg}(x) \\
    = & ~ (1-\gamma) (A_{*,i}^\top \cdot (f(x) (f(x) - b)^\top f(x) + \diag(f(x))(f(x)-b)) \\
    & ~ -\gamma (-\langle f(x), A_{*,i}\rangle L_{\ent}(x) - \langle h(x), A_{*,i}\rangle) \\
    & ~ + A_{*,i}^\top W^2 Ax
\end{align*}
where the first step follows from the definition of $L$, the second step comes from Lemma \ref{lem:gradient_ent}, Lemma \ref{lem:L_reg_gradient_hessian} and Lemma 5.6 in \cite{dls23}.

{\bf Proof of Part 2.}
It simply follows from applying the result of {\bf Part 1} to all the $d$ coordinates in $x \in \R^d$.

\end{proof}

\newpage

\section{Hessian}\label{sec:hessian}
In Section \ref{sec:hessian:log_f}, we give the Hessian of $\log f(x)$.
In Section \ref{sec:hessian:ent}, we give the Hessian of $L_{\ent}$.

\subsection{Hessian for \texorpdfstring{$\log f(x)$}{}}\label{sec:hessian:log_f}
\begin{lemma}[Hessian of $\log f(x)$]\label{lem:hessian_log_f}
Provided that the following condition holds
\begin{itemize}
    \item We define function $f : \R^d \rightarrow \R^n$  as Definition \ref{def:f}.
\end{itemize}
We have
\begin{itemize}
    \item Part 1. 
    \begin{align*}
        \frac{\d^2 \log f(x)}{ \d x_i^2} = (  \langle f(x), A_{*,i} \rangle^2 - \langle f(x), A_{*,i} \circ A_{*,i} \rangle ) \cdot {\bf 1}_n.
    \end{align*}
    \item Part 2. 
    \begin{align*}
        \frac{\d^2 \log f(x)}{ \d x_i \d x_j} = ( \langle f(x), A_{*,i} \rangle \cdot \langle f(x), A_{*,j} \rangle - \langle f(x), A_{*,i} \circ A_{*,j} \rangle ) \cdot {\bf 1}_n.
    \end{align*}
\end{itemize}
\end{lemma}
\begin{proof}

{\bf Proof Part 1.}

For every $i \in [n]$, we have
    \begin{align*}
        \frac{\d^2 \log f(x)}{ \d x_i^2} 
        = & ~ \frac{\d  }{\d x_i} ( \frac{\d \log f(x)}{\d x_i} ) \\
        = & ~  \frac{\d}{\d x_i}(-\langle f(x) , A_{*,i} \rangle \cdot {\bf 1}_n + A_{*,i})\\
        = & ~ -\frac{\d}{\d x_i}( \langle f(x) , A_{*,i} \rangle \cdot {\bf 1}_n )\\ 
        = & ~ -\frac{\d}{\d x_i}( \langle f(x) , A_{*,i} \rangle ) \cdot {\bf 1}_n \\ 
        = & ~ (  \langle f(x), A_{*,i} \rangle^2 - \langle f(x), A_{*,i} \circ A_{*,i} \rangle ) \cdot {\bf 1}_n,
    \end{align*}
where the 1st step is due to basic calculus, the 2nd step comes from {\bf Part 1} of Lemma~\ref{lem:gradient_log_f}, the 3rd step follows from $A_{*,i}$ is independent of variable $x$, the 4th step comes from simple algebra, and the last step is due to {\bf Part 3} of Lemma~\ref{lem:gradient_f}.

{\bf Proof Part 2.}
 For every $i \in [n]$, we have
    \begin{align*}
        \frac{\d^2 \log f(x)}{ \d x_i \d x_j} 
        = & ~ \frac{\d }{\d x_i} ( \frac{\d \log f(x) }{ \d x_j} ) \\
        = & ~  \frac{\d}{\d x_i}(-\langle f(x) , A_{*,j} \rangle \cdot {\bf 1}_n + A_{*,j})\\
        = & ~ - \frac{\d}{\d x_i}(\langle f(x) , A_{*,j} \rangle \cdot {\bf 1}_n )\\ 
        = & ~ - \frac{\d }{ \d x_i} (\langle f(x) , A_{*,j} \rangle ) \cdot {\bf 1}_n \\
        = & ~ (  \langle f(x), A_{*,i} \rangle \langle f(x), A_{*,j} \rangle -\langle f(x), A_{*,i} \circ A_{*,j} \rangle ) \cdot {\bf 1}_n
    \end{align*}
where the 1st step follows basic calculus, the 2nd step comes from {\bf Part 1} of Lemma \ref{lem:gradient_log_f}, the 3rd step is due to $A_{*,j}$ is independent of variable $x$, the 4th step comes from simple algebra and the last step is using {\bf Part 3} of Lemma~\ref{lem:gradient_f}.
\end{proof}

\subsection{Hessian for \texorpdfstring{$L_{\ent}(x)$}{}}\label{sec:hessian:ent}

\begin{lemma}[Hessian of $L_{\ent}(x)$]\label{lem:hessian_L_ent}
If the following condition holds
\begin{itemize}
    \item  We define  $f$ as Definition \ref{def:f}.
    \item  We define  $L_{\ent}$ as Definition \ref{def:L_ent}
\end{itemize}
We have
\begin{itemize}

    \item Part 1.   
    \begin{align*}
        \frac{\d L_{\ent}(x)}{\d x_i^2}  = - \langle f(x), A_{*,i} \circ A_{*,i} \rangle \cdot L_{\ent}(x) + \langle f(x), A_{*,i}\rangle^2 - \langle f(x) +h(x),  A_{*,i} \circ A_{*,i} \rangle 
    \end{align*}

    \item Part 2.
    \begin{align*}
        \frac{\d L_{\ent}(x)}{\d x_i \d x_j}  = & ~ - \langle f(x), A_{*,i} \circ A_{*,j} \rangle \cdot L_{\ent}(x) - \langle f(x), A_{*,j}\rangle \cdot \langle h(x), A_{*,i}\rangle \\
    & ~ +  \langle f(x) , A_{*,i} \rangle \cdot \langle f(x) + h(x), A_{*,j} \rangle - \langle f(x) +h(x),  A_{*,i} \circ A_{*,j} \rangle \\
    \end{align*}
    
    \item Part 3. We can rewrite the $d \times d$ Hessian matrix  as
    \begin{align*}
        \frac{\d L_{\ent}(x)}{\d x^2}
    = & ~ A^\top (2 L_{\ent} + 1) f(x) f(x)^\top A + A^\top ( f(x) h(x)^\top + h(x) f(x)^\top ) A \\
     & ~ - A^\top \diag( (L_{\ent}(x) + 1)f(x) + h(x) ) A
    \end{align*}

\end{itemize}
\end{lemma}
\begin{proof}

{\bf Proof of Part 1.}
We have
\begin{align*}
    \frac{\d^2 L_{\ent}(x)}{\d x_i^2} = & ~ -\frac{\d^2 \langle f(x), \log(f(x)) \rangle}{ \d x_i^2 } \\
    = & ~ -( \frac{\d}{\d x_i} ( \langle f(x), A_{*,i}\rangle \cdot L_{\ent}(x)) + \frac{\d}{\d x_i} \langle h(x), A_{*,i}\rangle ) \\
    = & ~  2\langle f(x), A_{*,i}\rangle^2 \cdot L_{\ent}(x)- \langle f(x), A_{*,i} \circ A_{*,i} \rangle \cdot L_{\ent}(x) + \langle f(x), A_{*,i}\rangle \cdot \langle h(x), A_{*,i}\rangle \\
     & ~ + \langle f(x) , A_{*,i} \rangle \cdot \langle f(x) + h(x), A_{*,i} \rangle - \langle f(x) +h(x),  A_{*,i} \circ A_{*,i} \rangle \\
    = & ~ 2\langle f(x), A_{*,i}\rangle^2 \cdot L_{\ent}(x) - \langle f(x), A_{*,i} \circ A_{*,i} \rangle \cdot L_{\ent}(x) \\
    & ~ + \langle f(x), A_{*,i}\rangle \cdot \langle f(x)+2h(x), A_{*,i}\rangle - \langle f(x) +h(x),  A_{*,i} \circ A_{*,i} \rangle 
\end{align*}
where the first step follows from Definition \ref{def:L_ent}, the second step comes from the chain rule, the third step comes from Part 2 of Lemma \ref{lem:gradient_ent} and Part 2 of Lemma \ref{lem:gradient_h} and the last step comes from simple algebra.

{\bf Proof of Part 2.}
We have
\begin{align*}
    \frac{\d L_{\ent}(x)}{\d x_i \d x_j}  = & ~ -( \frac{\d}{\d x_i} ( \langle f(x), A_{*,j}\rangle \cdot L_{\ent}(x)) + \frac{\d}{\d x_i} \langle h(x), A_{*,j}\rangle ) \\
    =& ~ 2 \langle f(x), A_{*,j}\rangle \cdot \langle f(x), A_{*,i}\rangle L_{\ent} (x) \\
    & ~- \langle f(x), A_{*,i} \circ A_{*,j} \rangle \cdot L_{\ent}(x) \\
    & ~ + \langle f(x), A_{*,j}\rangle \cdot \langle h(x), A_{*,i}\rangle + \langle f(x) , A_{*,i} \rangle \cdot \langle  h(x), A_{*,j} \rangle\\
    & ~ + \langle f(x) , A_{*,i} \rangle \cdot \langle f(x) , A_{*,j} \rangle \\
    & ~ - \langle f(x) +h(x),  A_{*,i} \circ A_{*,j} \rangle )
\end{align*}
where the 1st step is using the chain rule, the 2nd step comes from Part 3 of Lemma \ref{lem:gradient_ent} and Part 3 of Lemma \ref{lem:gradient_h} and the reason for the last step is simple algebra.

{\bf Proof of Part 3.}

We have 
\begin{align*}
    \frac{\d^2 L_{\ent}}{ \d x^2}
    = & ~ 2 A^\top L_{\ent}(x) f(x) f(x)^\top A   \\
     & ~ - A^\top L_{\ent}(x) \diag( f(x) ) A \\
     & + ~ A^\top ( f(x) h(x)^\top + h(x) f(x)^\top ) A \\
     & + ~ A^\top f(x) f(x)^\top A \\
     & ~ - A^\top \diag( f(x) + h(x) ) A
\end{align*}
where the proof comes from combining {\bf Part 1} and {\bf Part 2}.
\end{proof}

\newpage
\section{Hessian is PSD}\label{sec:hessian_psd}
Following Lemma \ref{sec:hessian:ent}, we write the Hessian of $L_{\ent}$ as $A^\top B(x) A$.
In Section~\ref{sec:hessian_psd:B}, we show the bound of matrix $B(x)$.
In Section~\ref{sec:hessian_psd:result}, we show the lower bound of the Hessian of our loss function $L(x) = (1-\gamma) \cdot L_{\exp}(x) - \gamma \cdot L_{\ent}(x) + L_{\reg}(x)$. 

\subsection{Bounding Matrices \texorpdfstring{$B(x)$}{}}\label{sec:hessian_psd:B}
We define $B(x)$ as follows
\begin{definition}\label{def:B}
We define $B(x)$ as follows
\begin{align*}
 B(x) := & ~ (2 L_{\ent}(x) +1) f(x) f(x)^\top \\
 & ~ + f(x) h(x)^\top + h(x) f(x)^\top \\
 & ~ -  \diag(  (L_{\ent}(x)+1)f(x) + h(x) )
\end{align*}
Further, we define
\begin{align*}
    B_{\rank}(x) := & ~ \underbrace{ (2 L_{\ent}(x) +1) f(x) f(x)^\top }_{ := B_{\rank}^1(x) } + \underbrace{ f(x) h(x)^\top + h(x) f(x)^\top }_{ := B_{\rank}^2(x) } \\
    B_{\diag}(x):= & ~ -  \diag(  (L_{\ent}(x)+1)f(x) + h(x) )
\end{align*}
\end{definition}

\begin{lemma}\label{lem:hessian_psd_ent}
If the following conditions hold
\begin{itemize}
    \item Let $L_{\ent}(x) \in (0, \log n]$.
\end{itemize}
Then, we have
\begin{itemize}
    \item Part 1.
    \begin{align*}
      f(x) f(x)^\top \preceq  B_{\rank}^1(x) \preceq (3 \log n) \cdot f(x) f(x)^\top
    \end{align*}
    \item Part 2.
    \begin{align*}
        - ( f(x) f(x)^\top + h(x) h(x)^\top ) \preceq B_{\rank}^2(x) \preceq f(x) f(x)^\top + h(x) h(x)^\top
    \end{align*}
    \item Part 3.
    \begin{align*}
      - ( (2\log n) \| f(x) \|_{\infty} + \| h(x) \|_{\infty} ) I_n \preceq  B_{\diag}(x) \preceq ( (2\log n) \| f(x) \|_{\infty} + \| h(x) \|_{\infty} )  I_n
    \end{align*}
    \item Part 4. 
    \begin{itemize}
        \item If $\| f(x) \|_{\infty} \leq 1$.
        \item $\| h(x) \|_{\infty} \leq 1$
        \item $f(x) f(x) \preceq I_n$
        \item $h(x) h(x) \preceq \log^2 n \cdot I_n$
        \item Then we have
        \begin{align*}
           - 10 \log^2 n \cdot I_n \preceq  B(x) \preceq 10 \log^2 n \cdot I_n
        \end{align*}
    \end{itemize}
\end{itemize}
\end{lemma}
\begin{proof}

{\bf Proof of Part 1.}

We have
\begin{align*}
1 \leq 2 L_{\ent}(x) + 1 \leq 3 \log n
\end{align*}

\end{proof}

\subsection{Lower bound on Hessian}\label{sec:hessian_psd:result}
The objective of this section is to give the proof of Lemma~\ref{lem:hessian_pd}.
\begin{lemma}\label{lem:hessian_pd}
If the following conditions hold
\begin{itemize}
    \item Given matrix $A \in \R^{n \times d}$.
    \item Let $\gamma \in (0,1)$ denote the trade-off parameter that balances the creativity and reality.
    \item  We define  $L_{\ent}(x)$ as Definition~\ref{def:L_ent}.
    \item  We define  $L_{\reg}(x)$ as Definition~\ref{def:L_reg}.
    \item  We define  $L_{\exp}(x)$ as Definition~\ref{def:L_exp}.
    \item  We define  $L(x) = (1-\gamma) \cdot L_{\exp}(x) - \gamma \cdot L_{\ent}(x) + L_{\reg}(x)$. 
    \item Let $A^\top ( B(x) + W^2) A^\top$ be the Hessian of $L(x)$
    \item Let $W = \diag(w) \in \R^{n \times n}$. We denote $W^2 \in \R^{n \times n}$ as the matrix that $i$-th diagonal entry is $w_{i}^2$.
    \item Let $l > 0$ denote a scalar.
\end{itemize}
Then, we have
\begin{itemize}
    \item Part 1. If all $i \in [n]$, $w_{i}^2 \geq (1-\gamma) \cdot 10 + \gamma \cdot 10   \log^2 n+  l/\sigma_{\min}(A)^2$, then  
    \begin{align*}
    \frac{\d^2 L}{\d x^2} \succeq l \cdot I_d
    \end{align*}
    \item Part 2  
    If all $i \in [n]$, $w_{i}^2 \geq (1-\gamma) \cdot 200  +  \gamma \cdot 200 \log^2 n + l/\sigma_{\min}(A)^2$, then  
    \begin{align*}
      W^2 \approx_{1/10} B(x) +W^2
    \end{align*}
\end{itemize}
\end{lemma}
\begin{proof}

We define $B(x), B_1(x), B_2(x)$ such that 
\begin{itemize}
    \item $(1-\gamma) B_1(x) - \gamma B_2(x) = B(x)$
    \item $ \frac{\d^2 L_{\exp}}{\d x^2} = A^\top B_1(x)A$
    \item $ \frac{\d^2 L_{\ent}}{\d x^2} = A^\top B_2(x)A$
\end{itemize}

We have 
\begin{align*}
    \frac{\d^2 L }{\d x^2} = & ~ (1-\gamma) \frac{\d^2 L_{\exp}}{\d x^2} - \gamma \frac{\d^2 L_{\ent}}{\d x^2} + \frac{\d^2 L_{\reg}}{\d x^2} \\
    =& ~ (1-\gamma) A^\top B_1(x) A - \gamma A^\top B_2(x) A  + A^\top W^2 A \\
    = & ~ A^\top ( (1-\gamma) B_1(x) - \gamma B_2(x) + W^2) A 
\end{align*}
where the first step follows from the definition of $L$, the second step follows from Lemma \ref{lem:L_reg_gradient_hessian}, Lemma \ref{lem:hessian_L_ent} and Lemma 5.13 of \cite{dls23} and the last step follows from simple algebra.

We define $D$ as the following 
\begin{align*}
     D := (1-\gamma) B_1(x) - \gamma B_2(x) + W^2
\end{align*}
Then, we have
\begin{align*}
   D \succeq & ~  -(1-\gamma) 4 I_n + \gamma (- 10 \log^2 n \cdot I_n ) + w^2_{\min} I_n \\
    =  & ~ (-4(1-\gamma)   - \gamma ( 10 \log^2 n ) + w^2_{\min}) I_n \\
    \succeq & ~ \frac{l}{\sigma_{\min}(A)^2}  I_n
\end{align*}
where the first step follows from Lemma \ref{lem:hessian_psd_ent} and Lemma 6.2 in \cite{dls23}, the second step follows from simple algebra and the last step follows from $w_{i}^2 \geq (1-\gamma) \cdot 10 + \gamma \cdot 10   \log^2 n+  l/\sigma_{\min}(A)^2$.

Since $D$ is positive definite, we have
\begin{align*}
    A^\top D A \succeq \sigma_{\min}(D) \cdot \sigma_{\min}(A)^2  I_d \succeq l \cdot I_d.
\end{align*}
Hence, the Hessian of $L$ is positive definite.

{\bf Proof of Part 2.}

Using {\bf Part 4} of Lemma~\ref{lem:hessian_psd_ent}, we have
\begin{align*}
   & ~ - 10 \log^2 n \cdot I_n \preceq  B_2(x) \preceq 10 \log^2 n \cdot I_n \\
   & ~ 10 \log^2 n \cdot I_n \succeq  -B_2(x) \succeq -10 \log^2 n \cdot I_n \\
\end{align*}

Using Lemma~6.2 in \cite{dls23}, we have
\begin{align*}
   - 4 I_n \preceq  B_1(x) \preceq  4 \cdot I_n 
\end{align*}

By definition of $B(x)$, we have
\begin{align*}
    - (1-\gamma) 4 I_n -  10 \log^2 n \cdot I_n\preceq  (1-\gamma) B_1(x) - \gamma B_2(x) \preceq  (1-\gamma) 4 \cdot I_n+ 10 \log^2 n \cdot I_n
\end{align*}

From assumption on $W$, we also have
\begin{align*}
W^2
\succeq & ~ (1-\gamma) \cdot 200 I_n +  \gamma \cdot 200 \log^2 n I_n \\
- W^2 \preceq & ~ -(1-\gamma) \cdot 200 I_n -  \gamma \cdot 200 \log^2 n I_n \\
\end{align*}

Combining the above three equations, we have
\begin{align*}
-\frac{1}{20} W^2 \preceq B(x) \preceq \frac{1}{20} W^2
\end{align*}

Thus,
\begin{align*}
      B(x) + W^2 \approx_{1/20} W^2
\end{align*}

Then, we have
\begin{align*}
    W^2 \approx_{1/10}W^2 + B(x)
\end{align*}
\end{proof}

\newpage
\section{Hessian is Lipschitz}\label{sec:hessian_lipschitz}
In this section, we provide the main result of the Lipschitz.

\begin{lemma}\label{lem:hessian_lipschitz}
Provided that the following conditions hold
\begin{itemize}
    \item We define $B(x) \in \R^{n \times n}$ as Definition~\ref{def:B} (only based on $L_{\ent}$).
    \item Let $R_f:=\beta^{-2} n^{1.5} \exp(3R^2)$
    \item Let $\beta \geq \exp(-R^2)$
\end{itemize}
Then, we have
\begin{itemize}
    \item 
    \begin{align*}
    \| H(x) - H(y) \| \leq n^2 \exp(40R^2) \| x - y \|_2
    \end{align*}
\end{itemize}
\end{lemma}
\begin{proof}
We have
\begin{align*}
   \| H(x) - H(y) \|
   \leq & ~ \| A \| \| B(x) - B(y) \| \cdot \| A \| \\
   \leq & ~ R^2 \| B(x) - B(y) \| \\
   \leq & ~ R^2 ( 2\| G_1 \| + \sum_{i=2}^7 \| G_i \| ) \\
   \leq & ~ R^2 \cdot 100 R^2 \sqrt{n} \| f(x) - f(y) \|_2 \\
   \leq & ~ 100 R^4 \sqrt{n} \cdot R_f \cdot \| x - y \|_2 \\
   \leq & ~ n^2 \exp(40R^2) \cdot \| x - y \|_2
\end{align*}
where the forth step follows from Lemma~\ref{lem:summary_G_i}, and the last step follows from choice of $R_f$ and $\beta$ lower bound.
\end{proof}

\begin{lemma}[Lemma 7.1 in \cite{dls23}]\label{lem:lipschitz_L_exp}
 Let $\beta \geq \exp(-R^2)$. We define $H(x)$  as Hessian of $L_{\exp}$, then we have
\begin{align*}
    \| H(x) - H(y) \| \leq \beta^{-2} n^{1.5} \exp(20R^2) \| x-y \|_2.
\end{align*}
\end{lemma}

\section{Lipschitz for Scalar Function and Upper Bound}\label{sec:lipschitz_scalor}
In Section~\ref{sec:hessian_lipschitz:ent}, we provide the Lipschitz bound for  $L_{\ent}(x)$. 
Section~\ref{sec:hessian_lipschitz:upper_log_f_inf}, we provide the upper bound on $\|\log(1/f(x))\|_\infty$.

\subsection{Lipschitz for Scalar Function \texorpdfstring{$L_{\ent}(x)$}{}}\label{sec:hessian_lipschitz:ent}

\begin{lemma}\label{lem:L_ent_lipschitz}
Provided that the following conditions hold
\begin{itemize}
    \item  We define  $L_{\ent}$ as Definition \ref{def:L_ent}.
    \item  We define  $h$  as Definition \ref{def:h}.
\end{itemize}
Afterward, we obtain
\begin{align*}
    | L_{\ent}(x) - L_{\ent}(y) | \leq \sqrt{n} \cdot \| h(x) - h(y) \|_2
\end{align*}
\end{lemma}
\begin{proof}
We have
\begin{align*}
| L_{\ent}(x) - L_{\ent}(y) |
= & ~ | \langle h(x) - h(y) , {\bf 1}_n \rangle | \\
\leq & ~ \sqrt{n} \cdot \| h(x) - h(y) \|_2
\end{align*}
where the 1st step follows from Definition of $L_{\ent}$, the 2nd step comes from Fact~\ref{fac:basic_vector_norm}.
\end{proof}

\subsection{Upper Bound on \texorpdfstring{$\| \log(1/f(x))\|_{\infty}$}{}}\label{sec:hessian_lipschitz:upper_log_f_inf}

\begin{lemma}\label{lem:bound_linf_log_f}
Provided that the following conditions are met
\begin{itemize}
    \item $\alpha(x) \geq \beta$
    \item Let $\beta \geq \exp(-R^2)$
    \item $\| A \| \leq R$
    \item $\| x \|_2 \leq R$
\end{itemize}
Then we have
\begin{itemize}
\item Part 1.
\begin{align*}
    1 \geq f(x)_i \geq  \exp(-2R^2)
\end{align*}
\item Part 2.
\begin{align*}
    1 \leq 1/f(x)_i \leq \exp(2 R^2)
\end{align*}
\item Part 3.
\begin{align*}
    \| \log(1/f(x)) \|_{\infty} \leq 2R^2
\end{align*}
\end{itemize}
\end{lemma}
\begin{proof}
The proofs here are very straightforward, so we omit the details here.
\end{proof}

\section{Lipschitz for Vector Functions}\label{sec:lipschitiz_vector}

In Section~\ref{sec:hessian_lipschitz:f}, we provide the Lipschitz bound for $f(x)$. In Section~\ref{sec:hessian_lipschitz:log_f}, we provide the Lipschitz bound for $\log f(x)$.
In Section~\ref{sec:hessian_lipschitz:h}, we provide the Lipschitz bound for $h(x)$.

\subsection{Lipschitz for Vector Function \texorpdfstring{$f(x)$}{}}\label{sec:hessian_lipschitz:f}

We state a tool from previous work \cite{dls23}.
\begin{lemma}[Lemma 7.2 in \cite{dls23}]
Provided that the following conditions are met
\begin{itemize}
    \item $\langle \exp(Ax), {\bf 1}_n \rangle \geq \beta$
    \item Let $R_f:=\beta^{-2} n^{1.5} \exp(3R^2)$
\end{itemize}
Then, we have
\begin{align*}
    \| f(x) - f(y) \|_2 \leq R_f \cdot \| x - y \|_2.
\end{align*}
\end{lemma}

\subsection{Lipschitz for Vector Function \texorpdfstring{$\log f(x)$}{}}\label{sec:hessian_lipschitz:log_f}

The goal of this section is to prove Lemma~\ref{lem:lipschitz_log_f}.
\begin{lemma}\label{lem:lipschitz_log_f}
Provided that the following conditions hold
\begin{itemize}
    \item $\alpha(x) \geq \beta$
    \item $\| A \| \leq R$
    \item $\| x \|_2 \leq R$
    \item $\| {\bf 1}_n - f(y) / f(x) \|_{\infty} \leq 0.1$
\end{itemize}
Then we have
\begin{itemize}
\item Part 1.
\begin{align*}
    \| \log(1/f(x)) - \log(1/f(y)) \|_2 \leq  \| f(x) - f(y ) \|_2
\end{align*}
\item Part 2.
\begin{align*}
    \| \log(f(x) ) - \log(f(y)) \|_2 \leq \| f(x) - f(y) \|_2.
\end{align*}
\end{itemize}
\end{lemma}
\begin{proof}

{\bf Proof of Part 1.}
We have
\begin{align*}
 \| \log(1/f(x)) - \log(1/f(y)) \|_2 = & ~ \| \log(f(y)/f(x)) \|_2 \\
 \leq & ~ \| f(x) - f(y) \|_2
\end{align*}
where the 1st step follows from the definition of log, the 2nd step comes from Fact~\ref{fac:log}.

{\bf Proof of Part 2.}
It directly follows from Part 1.
\end{proof}

\subsection{Lipschitz for Vector Function \texorpdfstring{$h(x)$}{}}\label{sec:hessian_lipschitz:h}

\begin{lemma}\label{lem:lipschitz_h}
Provided that the following conditions are met
\begin{itemize}
    \item We define $f$  as Definition~\ref{def:f}.
    \item We define $h$  as Definition~\ref{def:h}.
\end{itemize}
We obtain
\begin{align*}
    \| h(x) - h(y) \|_2 \leq 3R^2 \| f(x) - f(y) \|_2 
\end{align*}
\end{lemma}
\begin{proof}
We have
\begin{align*}
\| h(x) - h(y) \|_2
= & ~ \| f(x) \circ \log f(x) - f(y) \circ \log f(y) \|_2 \\
= & ~ \| f(x) \circ \log(1/ f(x)) - f(y) \circ \log (1/f(y)) \|_2 \\
\leq & ~ \| ( f(x) - f(y) ) \circ \log(1/f(x)) \|_2 + \| f(y) \circ ( \log(1/f(x)) - \log(1/f(y)) ) \|_2 \\
\leq & ~ \| f(x) - f(y) \|_2 \cdot \| \log(1/f(x)) \|_{\infty} + \| \log(1/f(x)) - \log(1/f(y)) \|_2 \cdot \| f(y) \|_{\infty} \\
\leq & ~ 2R^2 \| f(x) - f(y) \|_2 +\| \log(1/f(x)) - \log(1/f(y)) \|_2 \\
\leq &  ~2R^2 \| f(x) - f(y) \|_2 + | f(x) - f(y) |_2 \\
\leq & ~ 3R^2 \| f(x) - f(y) \|_2,
\end{align*}
where the 1st step comes from the definition of $h(x)$, the 2nd step follows from simple algebra, the 3rd step comes from triangle inequality, the 4th step follows from Fact~\ref{fac:basic_vector_norm}, the 5th step comes from Lemma~\ref{lem:bound_linf_log_f}, and the 6th step follows from Lemma~\ref{lem:lipschitz_log_f}, the last step comes from $R \geq 1$.
\end{proof}

\section{Lipschitz for Matrix Functions}\label{sec:lipschitz_matrix}

In Section~\ref{lem:hessian_lipschitz_summary} for all the seven steps. 
In Section~\ref{sec:hessian_lipschitz:ent_f_f}, we provide the Lipschitz bound for $L_{\ent}(x) f(x) f(x)^\top$.
In Section~\ref{sec:hessian_lipschitz:f_f}, we provide the Lipschitz bound for $f(x) f(x)^\top$.
In Section~\ref{sec:hessian_lipschitz:f_h}, we provide the Lipschitz bound for  $f(x) h(x)^\top$.
In Section~\ref{sec:hessian_lipschitz:h_f}, we provide the Lipschitz bound for  $h(x) f(x)^\top$.
In Section~\ref{sec:hessian_lipschitz:diag_ent_f}, we provide the Lipschitz bound for  $\diag(L_{\ent}(x)f(x))$.
In Section~\ref{sec:hessian_lipschitz:diag_f}, we provide the Lipschitz bound for  $\diag(f(x))$.
In Section~\ref{sec:hessian_lipschitz:diag_h}, we provide the Lipschitz bound for  $\diag(h(x))$.

\subsection{Summary of Several Steps}\label{lem:hessian_lipschitz_summary}

\begin{lemma}\label{lem:summary_G_i}
Provided that the following conditions hold
\begin{itemize}
    \item $G_1 = L_{\ent}(x) f(x) f(x)^\top - L_{\ent}(y) f(y) f(y)^\top$
    \item $G_2 = f(x) f(x)^\top - f(y) f(y)^\top$
    \item $G_3= f(x) h(x)^\top - f(y) h(y)^\top$
    \item $G_4 = h(x) f(x)^\top - h(y) f(y)^\top$
    \item  $G_5 = \diag( L_{\ent}(x)   f(x)  ) - \diag( L_{\ent}(y)   f(y)  )$
    \item  $G_6 = \diag(  f(x)  ) - \diag(   f(y)  )$
    \item $G_7 = \diag(  h(x)  ) - \diag(   h(y)  )$
    \item Let $R \geq 1$
\end{itemize}
Then we have
\begin{align*}
    \| G_i \| \leq 4 \sqrt{n} R^2 \| f(x) - f(y) \|_2
\end{align*}
\end{lemma}
\begin{proof}
The proof directly follows from applying from Lemma~\ref{lem:G_1}, Lemma~\ref{lem:G_2}, Lemma~\ref{lem:G_3}, Lemma~\ref{lem:G_4}, Lemma~\ref{lem:G_5}, Lemma~\ref{lem:G_6} and Lemma~\ref{lem:G_7} to each $G_i$ for $i \in [7]$.
\end{proof}

\subsection{Lipschitz for Matrix Function \texorpdfstring{$L_{\ent}(x) f(x) f(x)^\top$}{}}\label{sec:hessian_lipschitz:ent_f_f}
We show how to bound $G_1$ as follows.
\begin{lemma}\label{lem:G_1}
If the following condition holds
\begin{itemize}
    \item $G_1 = L_{\ent}(x) f(x) f(x)^\top - L_{\ent}(y) f(y) f(y)^\top$
\end{itemize}
We have
\begin{align*}
    \| G_1 \| \leq 4 \sqrt{n} R^2 \cdot \| f(x) - f(y) \|_2
\end{align*}
\end{lemma}
\begin{proof}

We define
\begin{align*}
G_{1,1} :=  L_{\ent}(x) f(x) f(x)^\top  - L_{\ent}(x) f(x) f(y)^\top  \\
G_{1,2} :=  L_{\ent}(x) f(x) f(y)^\top  - L_{\ent}(x) f(y) f(y)^\top \\
G_{1,3} :=  L_{\ent}(x) f(y) f(y)^\top  - L_{\ent}(y) f(y) f(y)^\top
\end{align*}

We have
\begin{align*}
\| G_{1,1} \|
\leq & ~ L_{\ent}(x) \cdot \| f(x) \|_2 \cdot \| f(x) - f(y) \|_2 \\
\leq & ~ \log n \cdot \| f(x) - f(y) \|_2
\end{align*}

We have
\begin{align*}
    \| G_{1,2} \| 
    \leq & ~ L_{\ent}(x) \cdot \| f(x) - f(y) \|_2 \cdot \| f(y) \|_2 \\
    \leq & ~ \log n \cdot \| f(x) - f(y) \|_2
\end{align*}

We have
\begin{align*}
\| G_{1,3} \|
\leq & ~ | L_{\ent}(x) - L_{\ent}(y) | \cdot \| f(y) f(y)^\top \| \\
\leq & ~ | L_{\ent}(x) - L_{\ent}(y) | \cdot \| f(y) \|_2^2 \\
\leq & ~ | L_{\ent}(x) - L_{\ent}(y) | \\
\leq & ~ \sqrt{n} \cdot \| h(x) - h(y) \|_2 \\
\leq & ~ \sqrt{n} \cdot 3 R^2 \| f(x) - f(y) \|_2
\end{align*}
where the first step comes from Fact~\ref{fac:basic_matrix_norm}, the second step follows from Fact~\ref{fac:basic_matrix_norm}, the third step comes from Fact~\ref{fac:f}, the forth step follows from Lemma~\ref{lem:L_ent_lipschitz}, and the last step comes from Lemma~\ref{lem:lipschitz_h}.

\end{proof}

\subsection{Lipschitz for Matrix Function \texorpdfstring{$f(x) f(x)^\top$}{}}\label{sec:hessian_lipschitz:f_f}
\begin{lemma}\label{lem:G_2}
Provided that the following conditions hold
\begin{itemize}
    \item $G_2 = f(x) f(x)^\top - f(y) f(y)^\top$
\end{itemize}
We have
\begin{align*}
    \| G_2 \| \leq 2 \| f(x) - f(y) \|_2
\end{align*}
\end{lemma}
\begin{proof}
We define 
\begin{align*}
G_{2,1} := f(x) f(x)^\top - f(x) f(y)^\top \\
G_{2,2} := f(x) f(y)^\top - f(y) f(y)^\top
\end{align*}
We can show 
\begin{align*}
    \| G_{2,1} \| \leq & ~ \| f(x) \|_2 \cdot \| f(x) - f(y) \|_2 \\
    \leq & ~ \| f(x) -f(y) \|_2
\end{align*}
Similarly, we can show
\begin{align*}
    \| G_{2,2} \| \leq & ~ \| f(x) - f(y) \|_2 \cdot \| f(y) \|_2 \\
    \leq & ~ \| f(x) - f(y) \|_2
\end{align*}

Thus, we have
\begin{align*}
\| G_2 \| \leq & ~ \| G_{2,1} \| + \| G_{2,2} \| \\
\leq & ~ 2 \| f(x) - f(y) \|_2
\end{align*}
Thus, we complete the proof.
\end{proof}

\subsection{Lipschitz for Matrix Function \texorpdfstring{$f(x) h(x)^\top$}{}}\label{sec:hessian_lipschitz:f_h}

\begin{lemma}\label{lem:G_3}
If the following condition holds
\begin{itemize}
    \item $G_3= f(x) h(x)^\top - f(y) h(y)^\top$
\end{itemize}
Then we have
\begin{align*}
    \| G_3 \| \leq 6R^2 \log n \cdot \| f(x) - f(y) \|_2.
\end{align*}
\end{lemma}
\begin{proof}
We define
\begin{align*}
    G_{3,1} := & ~ f(x) h(x)^\top - f(y) h(x)^\top \\
    G_{3,2} := & ~ f(y) h(x)^\top - f(y) h(y)^\top
\end{align*}
Then we have 
\begin{align*}
    \| G_{3,1} \| 
    \leq & ~ \| f(x) -f(y) \|_2 \cdot \| h(x) \|_2 \\
    \leq & ~ (\log n ) \cdot \| f(x) - f(y) \|_2
\end{align*}
where the second step comes form Fact~\ref{fac:h}.

We have
\begin{align*}
    \| G_{3,2} \|
    \leq & ~ \| f(y) \|_2 \cdot \| h(x) - h(y) \|_2 \\
    \leq & ~ 1 \cdot \| h(x) - h(y) \|_2 \\
    \leq & ~ 3R^2 \| f(x) - f(y) \|_2
\end{align*}
where the second step comes from Fact~\ref{fac:f}, and the last step comes from Lemma~\ref{lem:lipschitz_h}.

\end{proof}

\subsection{Lipschitz for Matrix Function \texorpdfstring{$h(x) f(x)^\top$}{}}\label{sec:hessian_lipschitz:h_f}

\begin{lemma}\label{lem:G_4}
If the following condition holds
\begin{itemize}
    \item $G_4 = h(x) f(x)^\top - h(y) f(y)^\top$
\end{itemize}
Then we have
\begin{align*}
    \| G_4 \| \leq 6 R^2 \log n \cdot \| f(x) - f(y) \|_2.
\end{align*}  
\end{lemma}
\begin{proof}
The proof is identical to Lemma~\ref{lem:G_3}. So we omitted the details here.
\end{proof}

\subsection{Lipschitz for Matrix Function \texorpdfstring{$\diag( L_{\ent}(x)   f(x)  )$}{}}\label{sec:hessian_lipschitz:diag_ent_f}

\begin{lemma}\label{lem:G_5}
If the following condition holds
\begin{itemize}
    \item $G_5 = \diag( L_{\ent}(x)   f(x)  ) - \diag( L_{\ent}(y)   f(y)  )$
\end{itemize}
Then we have
\begin{align*}
    \| G_5 \| \leq 6 \sqrt{n} R^2 \cdot \| f(x) - f(y) \|_2.
\end{align*}  
\end{lemma}
\begin{proof}
We define 
\begin{align*}
G_{5,1} = & ~ \diag( L_{\ent}(x) f(x) ) - \diag( L_{\ent}(x) f(y)  ) \\
G_{5,2} = & ~ \diag( L_{\ent}(x) f(y)  )  - \diag( L_{\ent}(y) f(y)  ) 
\end{align*}
We have
\begin{align*}
\| G_{5,1} \| \leq & ~ | L_{\ent(x)} | \cdot \| f(x) - f(y) \|_2  \\
\leq & ~ (\log n) \cdot \| f(x) - f(y) \|_2
\end{align*}
where the second step comes from Fact~\ref{fac:L_ent}.

We have
\begin{align*}
\| G_{5,2} \| 
\leq & ~ |L_{\ent}(x) - L_{\ent}(y)| \cdot \| f(y) \|_2 \\
\leq & ~ |L_{\ent}(x) - L_{\ent}(y)| \cdot 1 \\
\leq & ~ \sqrt{n} \cdot \| h(x) - h(y) \|_2 \\
\leq & ~ \sqrt{n} 3 R^2 \| f(x) - f(y) \|_2
\end{align*}
where the second step comes from Fact~\ref{fac:f}, the forth step comes from Lemma~\ref{lem:L_ent_lipschitz}, the last step comes from Lemma~\ref{lem:lipschitz_h}
\end{proof}

\subsection{Lipschitz for Matrix Function \texorpdfstring{$\diag( f(x)  )$}{}}\label{sec:hessian_lipschitz:diag_f}

\begin{lemma}\label{lem:G_6}
If the following condition holds
\begin{itemize}
    \item $G_6 = \diag(  f(x)  ) - \diag(   f(y)  )$
\end{itemize}
Then we have
\begin{align*}
    \| G_6 \| \leq  \| f(x) - f(y) \|_2.
\end{align*}  
\end{lemma}
\begin{proof}
We have
\begin{align*}
    \| G_6 \| 
    = & ~ \| \diag(f(x) ) -\diag(f(y)) \| \\
    \leq & ~ \| f(x) - f(y) \|_2
\end{align*}
\end{proof}

\subsection{Lipschitz for Matrix Function \texorpdfstring{$\diag(  h(x) )$}{}}\label{sec:hessian_lipschitz:diag_h}

\begin{lemma}\label{lem:G_7}
If the following condition holds
\begin{itemize}
    \item $G_7 = \diag(  h(x)  ) - \diag(   h(y)  )$
\end{itemize}
Then we have
\begin{align*}
    \| G_7 \| \leq 3R^2 \| f(x) - f(y) \|_2.
\end{align*}  
\end{lemma}
\begin{proof}
We can show
\begin{align*}
\| \diag(  h(x)  ) - \diag(   h(y)  ) \|
\leq & ~ \| h(x) - h (y) \|_2 \\
\leq & ~3R^2 \| f(x ) - f(y) \|_2
\end{align*}
where the last step comes from Lemma~\ref{sec:hessian_lipschitz:h}.
\end{proof}

\newpage
\section{Newton Step}\label{sec:newton}

In Section~\ref{sub:newton:good}, we present our loss function $L(y)$ and refer to it as being $(l, M)$-good when it meets some criteria. Moving on to Section~\ref{sub:newton:approximation}, we give the definition of  approximated Hessian and approximate update in Newton method.

\subsection{\texorpdfstring{$(l,M)$}{}-good Loss function}
\label{sub:newton:good}

We first provide the assumption of our loss function.

Given that $L(y) =(1-\gamma) \cdot L_{\mathrm{reality}}(y) + \gamma \cdot L_{\mathrm{creavity}}(y) + L_{\reg}(y)$ 
, we consider the following  optimization problem 
\begin{align*}
    \min_{x \in \R^d} L(y).
\end{align*}

We call the function $L$ $(l,M)$-good if it satisfies the assumption defined in Definition \ref{def:assumptions}.

\subsection{Approximated Hessian and Newton Update}\label{sub:newton:approximation}

Now, we give the definition of the approximate update. We first define the approximated Hessian as the following:

\begin{definition}[Approximated Hessian]\label{def:wt_H}
Let $\epsilon_0 \in (0,0.1)$ be a parameter.

Given Hessian $H(y_t) \in \R^{ d \times d}$, we consider the approximated Hessian $\wt{H}(y_t) \in \R^{d \times d}$ that it satisfies the following condition:
\begin{align*}
    \wt{H}(y_t) \approx_{\epsilon_0} H(y_t)
\end{align*}
\end{definition}

To efficiently obtain the approximate Hessian matrix $\wt{H}(y_t)$, we present a commonly used technique that makes use of the leverage score sampling (see Lemma~4.5 in \cite{dsw22}).
\begin{lemma}[\cite{dsw22,syyz22}]\label{lem:subsample}
Consider a constant precision parameter denoted by $\epsilon_0=0.01$. 
Consider the failure probability denoted by $\delta \in (0,1)$.
If we have a matrix $B \in \R^{n \times d}$, then, for any positive diagonal matrix $Q \in \R^{n \times n}$, we can find an algorithm that has the following time complexity 
\begin{align*}
O( (\nnz(B) + d^{\omega} ) \poly(\log(n/\delta)) ).
\end{align*}

This algorithm outputs an $O(d \log(n/\delta))$ sparse diagonal matrix $\wt{Q} \in \R^{n \times n}$, satisfying 
\begin{align*}
    B^\top \wt{Q} B \approx_{\epsilon_0}  B^\top Q B 
\end{align*}

\end{lemma}

\begin{definition}[Approximate update]\label{def:update_x_t}

We consider the following approximate update process
\begin{align*}
    y_{t+1} = y_t  - \wt{H}(y_t)^{-1} \cdot  g(y_t)  .
\end{align*}
\end{definition}

Then, we state a tool from prior work (Lemma 6.9 in \cite{lsz23}),
\begin{lemma}[\cite{lsz23}]\label{lem:one_step_shrinking}
Provided that the following conditions are met
\begin{itemize}
    \item Loss Function $L$ is $(l,M)$-good. (The definition of $(l,m)$-good can be found in   Definition~\ref{def:assumptions}). 
    \item Let $\epsilon_0 \in (0,0.1)$. 
    \item We define $y^*$ as Definition~\ref{def:assumptions}.
    \item We define $y_t$ as Definition~\ref{def:update_x_t}.
    \item Let $\mathrm{res}_t:= \| y_t - y^* \|_2$.
    \item Let $\ov{\mathrm{res}}_t: = M \cdot \mathrm{res}_t$
\end{itemize}
 Then we have  
\begin{align*}
\mathrm{res}_{t+1} \leq 2 \cdot (\epsilon_0 + \ov{\mathrm{res}}_t/( l - \ov{\mathrm{res}}_t ) ) \cdot \mathrm{res}_t.
\end{align*} 
\end{lemma}

In this context, $T$ symbolizes the overall count of iterations performed by the algorithm. In order to use Lemma~\ref{lem:one_step_shrinking}, it is necessary to introduce the following induction lemma. This lemma is commonly found in existing literature, specifically mentioned in Lemma 6.10 in \cite{lsz23}.
\begin{lemma}[\cite{lsz23}]\label{lem:newton_induction}
Provided that the following condition are met
\begin{itemize}
    \item For every $i \in [t]$, we define $\mathrm{res}_i:= \| y_i - y^* \|_2$
    \item We define $y_i$ as Definition~\ref{def:update_x_t}.
    \item We define $y^*$ as Definition~\ref{def:assumptions}.
    \item $\epsilon_0 = 0.01$ (Definition~\ref{def:wt_H} for $\epsilon_0$)
    \item $\mathrm{res}_{i} \leq 0.4 \cdot \mathrm{res}_{i-1}$, for every $i \in [t]$
    \item $M \cdot \mathrm{res}_i \leq 0.1 l$, for every $i \in [t]$
\end{itemize}
We obtain
\begin{itemize}
    \item $\mathrm{res}_{t+1} \leq 0.4 \mathrm{res}_t$
    \item $M \cdot \mathrm{res}_{t+1} \leq 0.1 l$
\end{itemize}
\end{lemma}

\section{Main Result}\label{sec:main_result}

\begin{theorem}[Restatement of Theorem~\ref{thm:main_informal}]\label{thm:main_formal}
Given matrix $A \in \R^{n \times d}$, $b \in \R^n$, and $w \in \R^n$. 
\begin{itemize}
    \item We define $f(x):=\langle \exp(Ax), \mathbf{1}_n \rangle^{-1} \exp(Ax)$.
    \item We use $x^*$ to denote the optimal solution of  
    \begin{align*}
    \min_{x \in \R^d} (1-\gamma) \cdot L_{\exp}(x) - \gamma \cdot L_{\ent}(x) + L_{\reg}(x)
    \end{align*}
    that
    \begin{itemize}
        \item $g(x^*) = {\bf 0}_d$, where $g$ denotes the gradient function.
        \item $\| x^* \|_2 \leq R$.
    \end{itemize}
    \item Suppose that $R \geq 10$.
    \item Assume that $\| A \| \leq R$. Here $\| A \|$ denotes the spectral norm of matrix $A$.
    
    \item Suppose that $b \geq {\bf 0}_{n}$ and $\| b \|_1 \leq 1$.\footnote{ Here ${\bf 0}_n$ denotes a length-$n$ vector where every coordinates are zeros. (Here $b \geq {\bf 0}_n$ denotes $b_i \geq 0$ for all $i\in[n]$). }

    \item Assume that $w_{i}^2 \geq 100 + l/\sigma_{\min}(A)^2$ for all $i \in [n]$.\footnote{We recall that $\sigma_{\min}(A)$ denotes the smallest singular value of matrix $A$.} 

    \item Let $M = n^{2} \exp(30 R^2)$. 
    \item Let $l > 0$.
    \item Let $x_0$ denote an starting/initial point such that $M \| x_0 - x^* \|_2 \leq 0.1 l$.
    \item We denote $\delta \in (0,0.1)$ and $\epsilon \in (0,0.1)$ as our failure probability and accuracy parameter respectively. 
    \item We denote $\omega$ as the exponent of matrix multiplication.
\end{itemize}
There is an algorithm (Algorithm~\ref{alg:main}) that
\begin{itemize}
\item  runs $\log(\| x_0 - x^* \|_2/ \epsilon)$ iterations 
\item spend  
\begin{align*}
O( (\nnz(A) + d^{\omega} ) \cdot \poly(\log(n/\delta)) 
\end{align*}
time per iteration,
\item and finally outputs a vector $\wt{x} \in \R^d$ such that
\begin{align*}
\Pr[ \| \wt{x} - x^* \|_2 \leq \epsilon ] \geq 1-\delta.
\end{align*}
\end{itemize}
\end{theorem}
\begin{proof}

Following from Lemma~\ref{lem:hessian_pd}, we know the Hessian of the loss function is positive definite.

Following from Lemma~\ref{lem:lipschitz_L_exp}, we know the Hessian of  $L_{\exp}$ is Lipschitz.

Following from Lemma~\ref{lem:hessian_lipschitz}, we know the Hessian of  $L_{\ent}$ is Lipschitz.

Then we run the algorithm in Newton section~\ref{sec:newton}, we complete the proof.

\end{proof}

\begin{algorithm}[!ht]\caption{Our Algorithm.
}\label{alg:main}
\begin{algorithmic}[1]
\Procedure{OurAlgorithm}{$A \in \R^{n \times d},b \in \R^n,w \in \R^n, \epsilon, \delta$} \Comment{Theorem~\ref{thm:main_formal}} 
    \State Initialize the value of $x_0$ (satisfying Definition~\ref{def:assumptions})  
    \State Initialize the number of iteration:  $T \gets \log( \| x_0 - x^* \|_2 / \epsilon )$.  
    \For{$t=0 \to T$} 
        \State $Q \gets B_{\diag}(x_t) + \diag(w \circ w)$ 
        \State $\wt{Q} \gets \textsc{SubSample}(Q,A,\epsilon_1 = \Theta(1), \delta_1 = \delta/T)$ \Comment{Lemma~\ref{lem:subsample}}
        \State Compute gradient $g$ exactly \Comment{Lemma~\ref{lem:gradient_loss_L}}
        \State $\wt{H} \gets A^\top \wt{Q} A$ 
        \State $x_{t+1} \gets x_t + \wt{H}^{-1} g$ 
        \Comment{Definition~\ref{def:update_x_t}}
    \EndFor
    \State $\wt{x}\gets x_{T+1}$
    \State \Return $\wt{x}$
\EndProcedure
\end{algorithmic}
\end{algorithm}

\ifdefined\isarxiv

\newpage
\bibliographystyle{alpha}
\bibliography{ref}

\else

\fi




\end{document}